\documentclass[11pt,a4paper,twoside]{scrartcl} %  report : scrreprt , article : scrartcl

\usepackage[english]{babel}
\usepackage[utf8]{inputenc}
\usepackage{graphicx}
\usepackage[dvipsnames]{xcolor}
\usepackage{cancel}
\usepackage{amsmath,amsthm}
\usepackage{amsfonts} 
\usepackage{amssymb,bbm}

\usepackage{gpeyre}
\usepackage{epstopdf}
\usepackage{epsfig}

\usepackage{url}
\usepackage{hyperref}
\hypersetup{
    hypertexnames=true,
    bookmarks=false,         % show bookmarks bar?
    unicode=false,          % non-Latin characters in Acrobat’s bookmarks
    pdftoolbar=true,        % show Acrobat’s toolbar?
    pdfmenubar=true,        % show Acrobat’s menu?
    pdffitwindow=false,     % window fit to page when opened
    pdfstartview={FitH},    % fits the width of the page to the window
    pdftitle={Convex Color Image Segmentation with Optimal Transport Distances},    % title
    pdfauthor={Julien Rabin and Nicolas Papadakis}; % {Rabin, Papadakis},     % author
    pdfsubject={Optimal Transport distance for image segmentation},   % subject of the document
    pdfcreator={Julien Rabin and Nicolas Papadakis},   % creator of the document
    pdfproducer={Julien Rabin and Nicolas Papadakis}, % producer of the document
    pdfkeywords={color segmentation} {optimal transport} {image processing}, % list of keywords
    pdfnewwindow=true,      % links in new window
    backref=true,               % Permet d'ajouter des liens dans
    pagebackref=true,       % les bibliographies     
    colorlinks=true,       % false: boxed links; true: colored links
    linkcolor=blue,          % color of internal links (change box color with linkbordercolor)
    citecolor=red,        % color of links to bibliography
    filecolor=magenta,      % color of file links
    urlcolor=blue,           % color of external links
}
\usepackage[hyperpageref]{backref} 

\usepackage{bclogo}
\usepackage{graphicx}
\graphicspath{{./images/}}

\newcommand{\U}{\mathbf{1}}
\renewcommand{\O}{\mathbf{0}}
\newcommand \Idc {\mathbbm{1}}

\newcommand{\R}{\RR}

\newcommand{\MK}{\textbf{MK}}

\newcommand{\A}{{\cal A}}
\newcommand{\B}{{\cal B}}
\newcommand{\C}{{\cal C}}
\newcommand{\F}{{\cal F}}
\renewcommand{\H}{{\cal H}}
\renewcommand{\L}{{\cal L}}
\renewcommand{\S}{{\cal S}}
\newcommand{\X}{{\cal X}}
\newcommand{\TV}{\mathop{TV}}

\newtheorem{remark}{Remark}
\newtheorem{proposition}{Proposition}
\newtheorem{corollary}{Corollary}

\title{Convex Color Image Segmentation\\with Optimal Transport Distances\footnote{A short version of this work has been published in the proceedings of SSVM'15.}}

\urldef{\mailjuju}\path|julien.rabin@unicaen.fr|
\urldef{\mailnico}\path|nicolas.papadakis@math.u-bordeaux.fr|

\author{Julien Rabin \textsuperscript{$1$} and Nicolas Papadakis \textsuperscript{$2$}\\[3mm]
\textsuperscript{$1$} GREYC, Université de Caen, CNRS UMR 6072, France\\
\mailjuju\\
\textsuperscript{$2$} CNRS, IMB, UMR 5251,
Université de Bordeaux, France\\
\mailnico\\}

\usepackage{fancyhdr}
\iffalse
\else
\fi

\begin{document}

\maketitle

\begin{abstract}
This work is about the use of regularized optimal-transport distances for convex, histogram-based image segmentation.
In the considered framework, fixed exemplar histograms define a prior on the statistical features of the two regions in competition. 
In this paper, we investigate the use of various transport-based cost functions as discrepancy measures and rely on a primal-dual algorithm to solve the obtained convex optimization problem.

\medskip

\end{abstract}

%%%%%%%%%%%%%%%%%%%%%%%%%%%%%%%%%%%%%%%%%%%%%%%%%%%
\section{Introduction}

\paragraph{Optimal transport}

Optimal transport theory has received a lot of attention during the last decade as it provides a powerful framework to address problems which embed statistical constraints.
Its successful application in various image processing tasks has demonstrated its practical interest (see \emph{e.g.}~\cite{bresson_wasserstein09,Papadakis_ip11,Rabin_icip11,Jungpeyre_wassestein11,TV_Kantorovich_SIAM14}).
Some limitations have been also shown and partially addressed, such as time complexity, regularity and relaxation~\cite{Cuturi13,Ferradans_siims14}.

\paragraph{Segmentation}

Statistical based image segmentation has been thoroughly studied in the literature, first using parametric models (such as the mean and variance), and then empirical distributions combined with adapted statistical distances, such as the Kullback-Leibler divergence.
In this work, we are interested in the use of the optimal transport framework for Image segmentation. This has been first investigated in~\cite{bresson_wasserstein09} for 1D features, then extended to multi-dimensional features using approximations of the optimal transport cost \cite{Jungpeyre_wassestein11,PeyreWassersteinSeg12}, and adapted to region-based active contour in \cite{PeyreWassersteinSeg12}, relying on a non-convex formulation.
In~\cite{Swoboda13}, a convex formulation is proposed, making use of sub-iterations to compute the proximity operator of the Wasserstein distance, which use is restricted to low dimensions.

In this paper, we extend the convex formulation for two-phase image segmentation of~\cite{papa_aujol} for non-regularized as well as regularized~\cite{Cuturi13,CuturiDoucet14} optimal transport distances.
This work shares some common features with the recent work of \cite{CuturiPeyreRolet_preprint} in which the authors
investigate the use of the Legendre-Fenchel transform of regularized transport cost for imaging problems.

\section{Convex histogram-based image segmentation}

\subsection{Notation}

We consider here vector spaces equipped with the scalar product $\dotp{.\,}{.}$ and the norm
$\lVert . \rVert = \sqrt{\dotp{.\,}{.}}$.
The conjugate operator of $A$ is denoted by $A^*$ and satisfies $\langle Ax,y \rangle = \langle x,A^* y \rangle$.
We denote as $\U_n \text{ and } \O_n \in \R^n$ the $n$-dimensional vectors full of ones and zeros respectively, $x^T$ the transpose of $x$, and $\nabla$ the discrete gradient operator, while
$\text{Id}$ stands for the identity operator.
The operator $\diag(x)$ defines a square matrix whose diagonal is $x$.
Functions $\iota_S$ and $\Idc_S$ are respectively the characteristic and indicator functions of a set $S$. 
$\textit{Proj}$ and $\textit{Prox}$ stands respectively for the Euclidean projection and proximity operator.
The set $\S_{k,n} := \{x \in \R^{n}_+, \dotp{x}{\U_n}=k\}$ is the simplex of histogram vectors ($\S_{1,n}$ being therefore the discrete probability simplex of $\R^n$).

\subsection{General formulation of distribution-based image segmentation}

Let $I : x \in \Om \mapsto I(x) \in \RR^{d}$ be a color image, defined over the $N$-pixel domain $\Om$ ($N = \lvert \Om\rvert$), and $\F$ a feature-transform of $n$-dimensional descriptors $\F I(x) \in \RR^n$.
We would like to define a binary segmentation $u:\Om \mapsto \{0,1\}$ of the whole image domain, using two fixed probability distributions of features $a$ and $b$.
Following the variational model introduced in \cite{papa_aujol}, we consider the energy
\eql{\label{eq:segmentation_energy}
J(u)  = \rho \TV(u) + D( a , h(u) ) + D( b , h(\U-u) )
}
where $\rho\ge 0$ is the regularization parameter, 
\begin{itemize}\setlength\itemsep{0.3em}
\item[\textbullet] the fidelity terms are defined using  $D ( .,. )$, a dissimilarity measure between features;
%= \lVert x-y \rVert_1$, which is the L1 norm (special case of $\MK$ cost when the ground cost function is uniform, except for identical bins: $C(i,j) = \frac{1}{2} (1- \delta(i-j))$);
\item[\textbullet] $h(u)$ is the empirical discrete probability distribution of features $\F I$ using the binary map~$u$, which is written as a sum of Dirac masses % (when not using a kernel estimator)
\eq{
h(u) : y \in \R^n \mapsto \frac{1}{\sum_{x \in \Om} u(x)} \sum_{x \in \Om} u(x) \delta_{\F I(x)}(y) \;;
}
\item[\textbullet] $\TV(u)$ is the total variation norm of the binary image $u$, which is related to the perimeter of the region $R_1(u) := \{x \in \Om \,\vert\, u(x)=1\}$ (co-area formula).
\end{itemize}
Observe that this energy is highly non-convex since $h$ is a non linear operator, 
and that we would like to find a minimum over the non-convex set $\{0,1\}^N$.

\subsection{Convex relaxation of histogram-based segmentation energy}

The authors of \cite{papa_aujol} propose some relaxations and a reformulation in order to handle the minimization of energy \eqref{eq:segmentation_energy} using convex optimization tools.

\subsubsection{Probability map} The first relaxation consists in using a segmentation variable $u:\Om \mapsto [0,1]$ which is a weight function (probability map). 
A threshold is therefore required to get a binary segmentation of the image into two regions
$R_t(u) := \{x \in \Om \,\vert\, u(x) \ge t\}$ and its complement $R_{t}(u)^c$.

\subsubsection{Feature histogram}
The feature histogram of the probability map is denoted $H_\X(u)$ and defined as the \textbf{quantized, non-normalized, and weighted histogram} of the feature image $\F I$ using the relaxed variable $u:\Om \mapsto [0,1]$ and a feature set $\X=\{X_i \in \R^n\}_{1\le i\le M_X}$ composed of $M_X$ bins
\eq{
\left(H_\X(u)\right)_i =  \sum_{x \in \Om} u(x) \Idc_{\C_\X(i)} (\F I(x)), 
\quad
\forall\, i \in \{1,\ldots M_X\}  
}
where $i$ a bin index, $X_i$ is the centroid of the corresponding bin, and $\C_\X(i) \subset \RR^n$ 
is the corresponding set of features (\emph{e.g.} the Voronoï cell obtained from \emph{hard assignment} method).
Therefore, we can write $H_\X$ as a linear operator
\eq{\label{eq:histogram_matrix_def}
H_\X : u \in \R^{N} \mapsto \Idc_{\X} \cdot u \in \R^{M_X},
\;\; \text{ with} \;
\Idc_{\X}(i,j) := 1 \text{ if } \F I(j) \in \C_\X(i), 0 \text{ otherwise}.
}
Note that $\Idc_\X\in \RR^{M_X \times N}$ is a fixed assignment matrix that indicates which pixels of $\F I$ contribute to each bin $i$ of the histogram.
As a consequence, 
$
	\dotp{H_\X (u)}{\U_{X}} % =\sum_{i=1}^{M_X}\sum_{j \in \Om} \Idc_\X(i,j) 
	  = \sum_{x \in \Om} u(x) = \dotp{u}{\U_N}
$, so that
$H_\X (u) \in \S_{M_X, \dotp{u}{\U}}$.

\subsubsection{Exemplar histograms}
The segmentation is driven from two fixed histograms $a \in \S_{M_a,1}$ and $b \in \S_{M_b,1}$, which are normalized (\emph{i.e.} sum to $1$), have respective dimension $M_a$ and $M_b$, and are obtained using the respective sets of features $\A$ and $\B$.
In order to measure the similarity between the non-normalized histogram $H_\A u$ and the normalized histogram $a$, while obtaining a convex formulation, we follow \cite{papa_aujol} and consider the fidelity term
$D \left( {a \dotp{u}{\U_N} , H_\A u} \right)$,
where the constant vector $a$ has been scaled to $H_\A u \in \S_{M_a, \dotp{u}{\U}}$.

\subsubsection{Segmentation energy}
Observe that the problem can now be written as finding the minimum of the following energy
\begin{equation*}
%\label{eq:convex_segmentation_energy0}
\begin{split}
\tilde E(u)  &= \rho \TV(u) 
+ \tfrac{1}{\gamma} D \left( {a \dotp{u}{\U_N} , H_\A u} \right) % \sum_{x \in \Om} u(x) 
+ \tfrac{1}{N-\gamma} D\left({b \dotp{\U_N-u}{\U_N}, H_\B (\U_N-u) } \right). % \sum_{x \in \Om} (1-u(x)) 
\end{split}
\end{equation*}
The constant $\gamma \in (0,N)$ is meant to compensate for the fact that 
the binary regions $R_t(u)$ and $R_t(u)^c$ may have different size. 
More precisely, as we are interested in a discrete probability segmentation map, we consider the following constrained problem:
\eq{%\vspace*{-2mm}
\min_{u \in [0,1]^{N}} \tilde E(u) = 
\min_{u \in \R^{N}} \left \{ E(u) := \; \tilde E(u) + \iota_{[0,1]^{N}}(u)
\right\}.
}

\subsubsection{Simplification}
From now on, and without loss of generality, we will assume that all histograms are computed using the same set of features, \emph{namely} $\A=\B$. We will also omit unnecessary subscripts in order to simplify notation. Moreover, we also omit the parameter $\gamma$ since its value seems not to be critical in practice, as demonstrated in \cite{papa_aujol}.
Finally, introducing linear operators 
\begin{equation}
	A := a  \U_{N}^T \in \R^{M \cdot N}
	\quad \text{ and }  \quad
	B := b \U_{N}^T \in \R^{M \cdot N}
\end{equation}
such that
$Au = (a  \U_{N}^T)u  = a \langle u, \U_{N} \rangle $, % = a  (\U_{N}^T . u)
and the usual discrete definition of total variation
$$
\TV(u) =  {\lVert {\nabla u }\rVert}_{1,2}  = \sum_{x \in \Om} {\lVert \nabla u(x) \rVert}_2
% =  \sum_{x \in \Om} \lVert \nabla u(x) \rVert_2
= \sum_{i=1}^N { \left (\sum_{j \sim i} {\lVert u_i - u_j \rVert}_2^2 \right) }^{\frac12}
$$
we have the following minimization problem:
\begin{equation}
	\label{eq:convex_segmentation_problem}
	\min_{u} \; % \in [0,1]^N
	\rho  {\lVert \nabla u\rVert} %\TV(u) 
	+  D(Au , H u)
	+  D(B(1-u) , H(1-u) )+ \iota_{[0,1]^{N}}(u).
\end{equation}
Notice that matrix $H \in\R^{M\cdot N}$ is sparse (with $N$ non zero values) and $A$ and $B$ are of rank $1$, so that storing or manipulating these matrices is not an issue.

In \cite{papa_aujol}, the distance function $D$ was defined as the $L_1$ norm.
In the following sections, we investigate the use of similarity measure based on optimal transport, which is known to be more robust and appropriate for histogram comparison. 
The next paragraph details the optimization framework used in this work.

\subsection{Optimization}
In order to solve \eqref{eq:convex_segmentation_problem}, we consider the following dualization of the problem using the Legendre-Fenchel transforms of the $L^2$ norm and the function $D$ 
\eql{\label{pb:dual_type}
\small
\begin{split}
	\hspace{-2mm}
	\min_{u \in \R^N}
	\max_{\substack{p_A^{},q_A^{},p_B^{},q_B^{} \in \R^M\\p^{}_C \in \R^{2N}}}
	 & \langle Hu,p_A^{}\rangle + \langle A^{}u,q_A^{}\rangle +\langle H(\U -u),p_B^{}\rangle +\langle B^{}(\U-u),q_B^{}\rangle  + \langle \nabla u,p^{}_C\rangle
	 \\[-3mm]
 	& %\hspace{-5mm}  
 	+ \iota_{[0,1]^{N}}(u)-D^*(p_A^{},q_A^{})-D^*(p_B^{},q_B^{})- \iota_{\lVert.\rVert\leq\rho}(p^{}_C),
 	 \\[-3mm]
\end{split}
}
where $\iota_{\lVert . \rVert \leq\rho}$ is the characteristic function of the convex $\ell_2$ ball of radius $\rho$,
while $D^*$ is the dual of the function $D$. In order to accommodate the different models studied in this paper, we assume here that $D^*$ is a sum of two convex functions $D^*=D_1^*+D_2^*$, where $D_1^*$ is non-smooth and $D_2^*$ is differentiable and  has a Lipschitz continuous gradient. 

We recover a general primal-dual problem of the form 
\begin{equation}\label{pb:primaldual}
\min_u\max_{p} \; \langle Ku,p\rangle+\iota_{[0,1]^{N}}(u) + H(u) - F^*(p)-G^*(p),
\end{equation}%
with primal variable $u\in \R^N$ and dual vector $p=[p_A^T,q_A^T,p_B^T,q_B^T,p_C^T]^T \in \R^{4M+2N}$, where
\begin{itemize}\setlength\itemsep{0.3em}
\item[\textbullet] $K=[H^T,A^T,-H^T,-B^T,\nabla^T]^T \in \R^{(4M+2N)\times N}$ is a sparse, linear operator;
%\\[1mm]

\item[\textbullet] $H$ is convex and smooth ( $H(u) = 0$ in the setting of problem \eqref{pb:primaldual}) with Lipschitz continuous gradient $\nabla H$ with constant $L_H$;

\item[\textbullet] $\iota_{[0,1]^{N}}(u)$ is convex and non-smooth;
%\\[1mm]

\item[\textbullet] $F^*(p)=D_1^*(p_A^{},q_A^{})+D_1^*(p_B^{},q_B^{}) + \iota_{\lVert . \rVert\leq\rho}(p^{}_C)$ is convex and non-smooth;
%\\[1mm]

\item[\textbullet] 
$G^*(p)=D_2^*(p_A^{},q_A^{})+D_2^*(p_B^{},q_B^{})- \dotp{H\U_{N}}{p_B^{}}- \dotp{B\U_{N}}{q_B^{}}$ is convex and differentiable with Lipschitz constant $L_{G^*}$.
\end{itemize}

To solve this problem, we consider the primal dual algorithm of~\cite{VuPrimalDual,CP14}
\begin{equation}\label{algo:primaldual}
\left\{
\begin{array}{ll}
u^{k+1}&=\text{Proj}_{[0,1]^{N}}\left(u^k-\tau (K^T p^k + \nabla H(u^k)) \right)\\[1mm]
p^{k+1}&=\text{Prox}_{\sigma F^*}\left(p^k+\sigma (K (2u^{k+1}-u^k)-\nabla G^*(p^k))\right)
\end{array}\right.
\end{equation}
that converges to a saddle point of \eqref{pb:primaldual} as soon as (see for instance~\cite[Eq. 20]{CP14})
\begin{equation}\label{time_step}\left(\tfrac1\tau  - {L_H}\right)\left(\tfrac1\sigma - L_{G^*}\right)\geq \lVert K \rVert^2.\end{equation}
Note this method may benefit from the recent framework proposed in \cite{Lorenz_Pock_inertial_FB_JMIV2015}, using
inertial acceleration and pre-conditionning.

%%%%%%%%%%%%%%%%%%%%%%%%%%%%%%%%%%%%%%%%%%%%%%%%%%%
\section{Monge-Kantorovitch distance for image segmentation}

\subsection{Wasserstein Distance and Optimal Transport problem}

\subsubsection{Optimal Transport problem}

We consider in this work the discrete formulation of the Monge-Kantorovitch optimal mass transportation problem (see \emph{e.g.}~\cite{Villani03}) between a pair of histograms $a \in \S_{M_a, k}$ and $b \in \S_{M_b, k}$. 
Given a fixed assignment cost matrix $C_{\A,\B} \in \R^{M_a \times M_b}$ between the corresponding histogram centroids $\A=\{A_i\}_{1\le i \le M_a}$ and $\B=\{B_j\}_{1\le j \le M_b}$, an optimal transport plan %$P^*\in {\Pp}(a,b)$ 
minimizes the global transport cost, defined as a weighted sum of assignments
\eql{\label{eq:OT}
\small
	{\forall \, (a,b) \in \S}, \quad
	\MK(a,b) 
		%= \dotp{P^{\star}}{C} 
		:= \min_{P \in {\Pp}(a,b)} \left\{\dotp{P}{C}  =  \sum_{i=1}^{M_a}\sum_{j=1}^{M_b}  P_{i,j}C_{i,j}\right\}.
}
The sets of admissible histogram and transport matrices are respectively
\eql{\label{eq:admissible_histograms}
	\S: = \{a \in \R^{M_a}, b \in \R^{M_b} \,\vert\, a>0, b>0  \text{ and } \dotp{a}{\U_{M_a}} = \dotp{b}{\U_{M_b}}\},
}
\eql{\label{eq:admissible_matrices}
	{\Pp}(a,b) := \{P\in \RR_+^{M_a \times M_b},  P\U_{M_b} = a \text{ and } P^T \U_{M_a}  = b\}.
}
Observe that the norm of histograms is not prescribed in $\S$, and that we only consider histograms with positive entries since null entries do not play any role.

\subsubsection{Wasserstein distance}
When using $C_{i,j} = \lVert A_i - B_j \rVert^{p}$, then $\textbf{W}_p(a,b) = \MK(a,b)^{1/p}$ is a metric between normalized histograms.
In the general case where $C$ does not verify such a condition, by a slight abuse of terminology we will refer to the $\MK$ transport cost function as the Monge-Kantorovich \emph{distance}.

\subsubsection{Monge-Kantorovich distance}
In the following, due to the use of duality, it would be more convenient to introduce the following reformulation:
\eql{\label{eq:OT_indicators}
	\forall\, a,b \qquad \MK(a,b) = \min_{P \in {\Pp}(a,b)} \dotp{P}{C}  
	+\iota_{\S}(a,b).
}

\subsubsection{LP formulation}
We can rewrite the optimal transport problem as a linear program (LP) with vector variables.
The primal and dual problems write
\eql{\label{eq:primal_dual_MK}
	\MK(\alpha) 
	= \min_{\substack{p\in \RR^{M_a\cdot M_b}\\\text{s.t. } p \ge 0, \; L^T p = \alpha}} 
	\dotp{c}{p} + \iota_{\S}(\alpha)
	\;
	= \max_{\substack{\beta \in \R^{M_a+M_b}\\\text{s.t. } L\beta \le c}}  \;
	\dotp{\alpha}{\beta}
	 .
}
where 
$\alpha$ is the concatenation of histograms: $\alpha^T = [a^T,b^T]$ and
the unknown vector $p \in \RR^{M_a\cdot M_b}$ corresponds to the bi-stochastic matrix $P$ being read column-wise (\emph{i.e.} $P_{i,j} = p_{i+(j-1)\cdot M_a}$).
The  $M_a + M_b$ linear marginal constraints on $p$ are defined by the matrix
$L^T \in \RR^{(M_a + M_b) \times (M_a M_b)}$ through equation $L^T p = \alpha$, where
\eq{
L^T= 
\begin{bmatrix}
\U_{M_b} e_1^T & \U_{M_b} e_2^T & \cdots & \U_{M_b} e_{M_a}^T\\
\text{Id}_{M_b}  & \text{Id}_{M_b}   & \cdots & \text{Id}_{M_b}  \\
\end{bmatrix}
\; \text{ with } \;
e_i(j) = \delta_{i-j} \;\; \forall\; j\le M_b
.
}
Note that we have the following property:
$(L\alpha)_{i,j} = 
\left(L
\tiny{ 
	\begin{bmatrix}
		a\\b
	\end{bmatrix}
}
\right)_{i,j}=a_i + b_j.$

The dual formulation shows that the function $\MK(\alpha)$ is not strictly convex in $\alpha$.
We draw the reader's attention to the fact that the indicator of set $\S$ is not required anymore with the dual formulation, which will later come in handy.

\subsubsection{Dual distance} From Eq.~\eqref{eq:primal_dual_MK}, we have that the Legendre–Fenchel conjugate of $\MK$ writes simply as the characteristic function of the set $\L_c := \{\beta \,\vert\, L\beta - c \le 0\}$ 
\eql{\label{eq:MK_dual}
\forall \beta \in \R^{M_a+M_b}, \quad
	\MK^*{\left( \beta \right)}= \iota_{L\beta\leq c}(\beta).
}

\subsection{Integration in the segmentation framework}
\label{sec:bidualization}

We propose to substitute in problem \eqref{eq:convex_segmentation_problem}
the dissimilarity functions by the convex Monge-Kantorovich optimal transport cost \eqref{eq:OT_indicators}.

In order to apply our minimization scheme described in \eqref{algo:primaldual}, we should be able to compute the proximity operator of $\MK^*$, which is the projection onto the convex set $\L_c$.
However, because the linear operator $L$ is not invertible, we cannot compute this projector in a closed form and  an optimization problem should be solved  at each iteration of the process \eqref{algo:primaldual} as in \cite{Swoboda13}.

\subsubsection{Bidualization}
To circumvent this problem, we resort to a bidualization 
to rewrite the $\MK$ distance as a primal-dual problem. First, we have that
$\MK^*( \beta ) = f^*(L\beta)$ with $f^*(r) = \iota_{r\le c}(r)$, so that $f(r) = \dotp{r}{c} + \iota_{r\ge 0}(r)$.
Then,
\eql{\label{eq:primal_dual_MK_splitting}
\begin{split}
\hspace*{-8mm}
	\MK^*(\beta) &
	= f^*( L\beta ) = \max_r\; \dotp{r}{L\beta} - f(r) = \max_r\; \dotp{r}{L\beta-c} - \iota_{\cdot\ge 0}(r)
	\\
	\MK(\alpha) &
	= \max_{\beta}\; \dotp{\alpha}{\beta}  - f^*( L\beta )
	= \max_{\beta}\; \dotp{\alpha}{\beta} + \min_r \dotp{r}{c-L\beta} + \iota_{\cdot \ge 0}(r)
	%\\
	%& 
	%= \max_{\beta} \; \min_{p \in \R^{M_a \cdot M_b}} \; \dotp{\alpha}{\beta} - \dotp{p}{L\beta} + f(p)
	\\
	& = \min_{r} \; \max_{\beta} \; 
	\dotp{r}{c} + \iota_{\cdot \ge 0}(r) + \dotp{\alpha - L^T r}{\beta}.
	\\
%	& =
%	\\
\end{split}
}

\subsubsection{Segmentation problem} Plugging the previous expression into Eq.~\eqref{pb:dual_type} enables us to solve it using algorithm~\eqref{algo:primaldual}.
Indeed, introducing new primal variables $r_A,r_B \in \R^{M^2}$ related to transport mapping, we recover the following primal dual problem
\eql{\label{pb:dual_type_MK}
\small
\begin{split}
	\hspace{-4mm}
	\min_{\substack{u \in \R^N\\r_A,r_B \in \R^{M^2}}}
	\max_{\substack{p_A^{},q_A^{},p_B^{},q_B^{} \in \R^M\\p^{}_C \in \R^{2N}}}
	 & \langle Hu,p_A^{}\rangle + \langle A^{}u,q_A^{}\rangle +\langle H(\U -u),p_B^{}\rangle +\langle B^{}(\U-u),q_B^{}\rangle
	 \\[-4mm]
 	& %\hspace{-5mm}  
 	\dotp{r_A}{c-L{\begin{bmatrix}p_A\\q_A\end{bmatrix}}} + % \tiny
 	\dotp{r_B}{c-L{\begin{bmatrix}p_B\\q_B\end{bmatrix}}}   + \langle \nabla u,p^{}_C\rangle
 	%- \dotp{L^T r_B}{[p_B^T,  q_B^T]^T}
 	 \\%[-3mm]
 	 & + \iota_{[0,1]^{N}}(u) + \iota_{\cdot \ge 0}(r_A) + \iota_{\cdot \ge 0}(r_B) - \iota_{\lVert.\rVert\leq\rho}(p^{}_C).
\end{split}
}
Observe that now we have a linear term $H(u,r_A,r_B) = \dotp{r_A+r_B}{c}$ whose gradient has a Lipschitz constant $L_H = 0$. We have also gained extra non smooth characteristic functions $\iota_{\cdot \ge 0}$, whose proximity operators are trivial (projection onto the positive quadrant $\R^{M^2}_+$: $\text{prox}_{\iota_{\ge 0}}(x) = \max\{\O,x\}$).

\subsubsection{Advantages and drawback} The main advantage of this new segmentation framework is that it makes use of optimal transport to compare histograms of features, 
without sub-iterative routines such as solving optimal transport problems to compute sub-gradients or proximity operators (see for instance~\cite{Cuturi13,Swoboda13}), %  [Jalal]
or without making use of approximation (such as the Sliced-Wasserstein distance~\cite{PeyreWassersteinSeg12}, generalized cumulative histograms~\cite{Papadakis_ip11} or entropy-based regularization~\cite{CuturiDoucet14}).
Last, the proposed framework is not restricted to Wasserstein distances, since it enables the use of any cost matrix, and does not depend on features dimensionality.

However, a major drawback of this method is that it requires two additional primal variables $r_A$ and $r_B$ whose dimension is $M^2$ in our simplified setting, $M$ being the dimension of histograms involved in the model. As soon as $M^2 \gg N$, the number of pixels, the proposed method could be significantly slower than when using $L^1$ as in \cite{papa_aujol} due to time complexity and memory limitation. This is more likely to happen when considering high dimensional features, such as patches or computer vision descriptors, as $M$ increases with feature dimension~$n$.

%%%%%%%%%%%%%%%%%%%%%%%%%%%%%%%%%%%%%%%%%%%%%%%%%%%
\section{Regularized $\MK$ distance for image segmentation}

As already mentioned in the last section, the previous approach based on optimal transport may be very slow for large histograms.
In such a case, we propose to use instead the entropy smoothing of optimal transport recently proposed and investigated in~\cite{Cuturi13,CuturiDoucet14,CuturiPeyreRolet_preprint}, that may offer increased robustness to outliers~\cite{Cuturi13}.
While it has been initially studied for probability simplex $\S_1$, we here investigate its use for our framework with unnormalized histograms on $\S$.

\subsection{Sinkhorn distances $\MK_{\lambda}$}

The entropy-regularized optimal transport problem \eqref{eq:OT_indicators} on set $\S$ (Eq.~\eqref{eq:admissible_histograms})~is
\eql{\label{eq:sinkhorn_distance_nonnormalized}
\begin{split}
	\MK_\lambda (a,b) 
	& %:=  \dotp{P_\lambda^{\star}}{C} 
	:= \min_{P \in {\Pp}(a,b)} \left\{\dotp{P}{C}  - \tfrac{1}{\lambda} h(P) \right\} + {\iota_{\S}(a,b)},
%	\\
%	\MK_\lambda (\alpha) 
%	& :=  \min_{\substack{p\in \RR^{M_a\cdot M_b}\\\text{s.t. } p \ge 0, \; L^T p = \alpha}}  \dotp{p}{c + \tfrac{1}{\lambda} \log p} + {\iota_{\S}(\alpha)}.
\end{split}
} where the entropy of the matrix $P$ is defined as $h(P) := -\dotp{P}{\log P}$.
% = - \sum_{i,j} P_{i,j} \log P_{i,j}.$
%
Thanks to the negative entropy term which is strictly convex, the regularized optimal transport problem has a unique minimizer, denoted $P_\lambda^{\star}$, which can be recovered using a fixed point algorithm studied by Sinkhorn (see \emph{e.g.} \cite{Cuturi13}). %\cite{sinkhorn}.
The regularized transport cost $\MK_\lambda (a,b)$ is thus referred to as the \emph{Sinkhorn distance}.

\subsubsection{Interpretation}
Another way to express the negative entropic term is:
\eq{
 - h(p) : p \in \R^{k}_+ \mapsto \textbf{KL}(p \Vert \U_k) \in \R, 
 \quad \text{ with } k = M_a\cdot M_b
}
that is the Kullback-Leibler divergence between transport map $p$ and the uniform mapping. 
This shows that, as $\lambda$ decreases, the model encourages smooth, uniform transport so that the mass is spread everywhere.
This also explains why this distance shows better robustness to outliers, as reported in \cite{Cuturi13}.
To conclude, one thus would like to use in practice large values of $\lambda$ to be close to the original Monge-Kantorovich distance, but low enough to deal with feature perturbation.

\subsubsection{Structure of the solution} 
First, the Sinkhorn distance \eqref{eq:sinkhorn_distance_nonnormalized} reads as
\eql{\label{eq:sinkhorn_distance_nonnormalized_bis}
\begin{split}
	\MK_\lambda (\alpha) 
	& %=\dotp{p^\star_\lambda}{c}
	:= \min_{\substack{p\in \RR^{M_a\cdot M_b}\\\text{s.t. } p \ge 0, \; L^T p = \alpha}}  \dotp{p}{c + \tfrac{1}{\lambda} \log p} + {\iota_{\S}(\alpha)}.
\end{split}
}
As demonstrated in~\cite{Cuturi13}, when writing the Lagrangian of this problem %\eqref{eq:sinkhorn_distance_nonnormalized}
with a multiplier $\beta$ to take into account the constraint $L^Tp=\alpha$, we can show that the respective solutions $p_\lambda^{\star}$ and $P_\lambda^{\star}$ of problem \eqref{eq:sinkhorn_distance_nonnormalized} and \eqref{eq:sinkhorn_distance_nonnormalized_bis} write
\eq{
\log p_\lambda^{\star} = \lambda (L\beta - c) - \U
\Leftrightarrow
(\log P_\lambda^{\star})_{i,j} = \lambda (u_i + v_i - C_{i,j}) - 1
\text{ with }
\beta = 
\begin{bmatrix}
u\\v
\end{bmatrix}
.
%\quad \text{that is } P^\lambda = diag(u) e^{-\lambda C} diag(v)
}

\begin{remark}
The constant $-1$ is due to the fact that we use the unnormalized KL divergence $\textbf{KL}(p \Vert \U_k)$, instead of $\textbf{KL}(p \Vert \frac{1}{k}\U_k)$ for instance.
\end{remark}

\subsubsection{Sinkhorn algorithm} 
Sinkhorn showed that the alternate normalization of rows and columns of any positive matrix $M$ converges to a unique bistochastic matrix $P= \diag(x) M \diag(y)$.  The following fixed-point iteration algorithm can thus be used to find the solution $P_\lambda^{\star}$: setting $M_\lambda = e^{-\lambda C}$, one has
\eq{%\hspace*{-5mm}
P_\lambda^{\star} = \diag(x^\infty) M_\lambda \diag(y^\infty)
\quad \text{ where }
x^{k+1} = \frac {a} {M_\lambda\, y^{k}} 
\; \text{ and } \;
y^{k+1} = \frac {b} {M_\lambda^T\, x^{k}} 
,
}
where $a$ and $b$ are the desired marginals of the matrix.
This result enables us to design fast algorithms to compute  the regularized optimal transport plan, and the the Sinkhorn distance or its derivative, as demonstrated in~\cite{Cuturi13,CuturiDoucet14}.

\subsection{Legendre–Fenchel transformation of Sinkhorn distance $\MK_\lambda$}

Now, in order to use the Sinkhorn distance in algorithm \eqref{algo:primaldual}, we need to compute its Legendre-Fenchel transform, which has been expressed in~\cite{CuturiDoucet14}.
\begin{proposition}[Cuturi-Doucet]
The convex conjugate of $\MK_\lambda (\alpha)$ reads
\eql{\label{eq:dual_sinkhorn_nonnormalized}
\MK^*_\lambda (\beta) 
=\tfrac{1}\lambda \left \langle  Q_\lambda(\beta) ,  \U  \right\rangle
\quad \text{ with }
Q_\lambda(\beta) := e^{\lambda (L\beta - c)-\U}.
}
\end{proposition}
We obtain a simple expression of the Legendre–Fenchel transform which is $C^\infty$, but unfortunately, its gradient is not Lipschitz continuous.

To overcome this problem, we propose two solutions in the next paragraphs: either we use a new normalized Sinkhorn distance (§~\ref{sec:normalized_sinkhorn_distance}), whose gradient is Lipschitz continuous (§~\ref{sec:normalized_sinkhorn_gradient}), or we rely on the use of proximity operator (§~\ref{sec:prox_sinkhorn_distance}).

\subsection{Normalized Sinkhorn distance $\MK_{\lambda,\le N}$ on $\S_{\le N}$}
\label{sec:normalized_sinkhorn_distance}

As the set $\S$ of admissible histograms does not prescribe the sum of histograms, we
consider here a different setting in which the histograms' total mass are bounded above by $N$, the number of pixels of the image domain $\Om$
\eql{\label{eq:set_normalized_histogram}
	\S_{\le N}: = \left\{a \in \R^{M_a}, b \in \R^{M_b} \,\Big\vert\, a>0, b>0, \dotp{a}{\U_{M_a}} = \dotp{b}{\U_{M_b}} \le N \right\}.
}
Moreover, as the transport matrix $P_\lambda^\star$ is not normalized (\emph{i.e.} $\dotp{P_\lambda^\star}{\U}\le N$), we also propose to use a slightly normalized variant of the entropic regularization: % normalization of the entropy term.
\eql{\label{eq:normalized_entropy}
 \tilde h(p) := N h\left(\tfrac{p}{N}\right) = - N \,\textbf{KL}\left(\tfrac{p}{N} \Vert \U\right) 
 = - \dotp{p}{\log p} + \dotp{p}{\U} \log N
 .
}
%This enables to have the following
\begin{corollary}
\label{corollary:dual_simplexNmax}
The convex conjugate of the normalized Sinkhorn distance
\eql{\label{eq:sinkhorn_distanceN}
\MK_{\lambda,\le N} (\alpha) :=  
	\min_{\substack{p\in \RR^{M_a\cdot M_b}\\\text{s.t. } p \ge 0, \; L^T p = \alpha}} 
	 \left\{\dotp{p}{c + \tfrac{1}{\lambda}\log p - \tfrac{\log N}{\lambda} \U} \right\} 
	 +{\iota_{\S_{\le N}}(\alpha)}
	 %
%\vspace*{-1.5mm}
} 
reads, using the matrix-valued function $Q_\lambda(.) \mapsto e^{\lambda (L. - c)-\U}$ defined in~\eqref{eq:dual_sinkhorn_nonnormalized}
\eql{\label{eq:dual_sinkhorn_inegality}
\MK^*_{\lambda,\le N} (\beta) = 
\left\{\begin{array}{ll}
\frac{N}{\lambda} \langle Q_\lambda(\beta), \U\rangle 
&\quad \text{ if } \langle Q_\lambda(\beta), \U\rangle \leq 1  % \text{otherwise.}
\\[3mm]
\frac{ N}{\lambda} \log \langle Q_\lambda(\beta), \U\rangle  + \frac{ N}{\lambda}
& \quad\text{ if }\langle Q_\lambda(\beta), \U\rangle \geq 1 
 \\
\end{array}\right.
}
\end{corollary}
\begin{proof}
See appendix \ref{proof_coro}. %A.1.
\end{proof}
Observe that the dual function $\MK^*_{\lambda,\le N}(\beta)$ is continuous for $\dotp{Q_\lambda(\beta^\star)}{\U}=1$.
Note also that the optimal matrix now is written
$P^\star_\lambda = N Q_\lambda(\beta^\star)$ 
if $\dotp{Q_\lambda(\beta^\star)}{\U} \le 1$,
and $P^\star_\lambda = N \tfrac{Q_\lambda(\beta^\star)}{\dotp{Q_\lambda(\beta^\star)}{\U}}$ otherwise.

\subsection{Gradient of $\MK_{\lambda,\le N}^*$}
\label{sec:normalized_sinkhorn_gradient}

From Corollary \ref{corollary:dual_simplexNmax}, we can express the gradient of $\MK_{\lambda,\le N}^*$ which is continuous
(writing $Q$ in place of $Q_\lambda(\beta)$ to simplify expression)
\eql{\label{eq:dual_derivative_inegality}
\nabla \MK^*_{\lambda,\le N} (\beta) = 
\left\{\begin{array}{cll}
	N & \left(  Q \U_{M_b}, \U_{M_a}^T Q  \right) 
	& \quad \text{ if } \langle Q , \U\rangle \leq  1  % \text{otherwise.}
\\[3mm]
	\frac	{N}{\langle Q , \U\rangle} & \left( Q \U_{M_b}, \U_{M_a}^T Q  \right)   
	& \quad \text{ if } \langle Q, \U\rangle \geq  1 \\
\end{array}\right.
.
}
We emphasis here that we retrieve a similar expression than the one originatively demonstrated in \cite{CuturiPeyreRolet_preprint}, where the authors consider the Sinkhorn distance on the probability simplex $\S_1$ (\emph{i.e.} the special case where $N=1$ and $\langle Q, \U\rangle =  1$).

\begin{proposition}\label{prop:lipschitz_dual}
The gradient $\nabla \MK^{*}_{\lambda,\le N}$ is a Lipschitz continuous function of constant $L_{\MK^*}$ bounded by  $2\lambda N$.
\end{proposition}
\begin{proof}
See appendix \ref{sec:proof_lipschitz_dual}.
\end{proof}

\subsection{Optimization using $\nabla \MK^*_{\lambda, \le N}$}

The general final problem we want to solve can be expressed as:
{\small
\eql{\label{eq:OT_segmentation_energy_reg}
\min_u  \rho \TV(u) 
+ \MK_{\lambda, \le N}{\left( H_a u, Au \right)}
+ \MK_{\lambda, \le N}{\left( H_b (\U -u), B(\U-u) \right)}+\iota_{[0,1]^N}(u).
%\vspace{-0.2cm}
}
}

\noindent
Using the Legendre–Fenchel transform, the problem \eqref{eq:OT_segmentation_energy_reg} can be reformulated as:
{\small
\begin{equation*}\label{pb:dual}
%\hspace*{-6mm}
\begin{split}
\min_u\max_{\substack{p_A^{},q_A^{}\\p_B^{},q_B^{},p^{}_C}}\hspace{-1.2pt}&\langle H^{}_au,p_A^{}\rangle + \langle A^{}u,q_A^{}\rangle +\langle H_b^{}(\U_{} -u),p_B^{}\rangle +\langle B^{}(\U_{} -u),q_B^{}\rangle + \langle \nabla u,p^{}_C\rangle\\ &+\iota_{[0,1]^{N}}(u)-\MK^*_{\lambda,\leq N}(p_A^{},q_A^{})-\MK^*_{\lambda,\leq N}(p_B^{},q_B^{})- \iota_{\lVert . \rVert\leq\rho}(p^{}_C),%\vspace{-0.1cm}
\end{split}
\end{equation*}
}

\noindent
and can be optimized with the algorithm \eqref{algo:primaldual}.
Using proposition \ref{prop:lipschitz_dual}, $\nabla G^*$ is a Lipschitz continuous function with constant $L_{G^*}$ checking
$
L_{G^*} = 2 L_{\MK^*} +\lVert H_b \rVert+\lVert B \rVert=2\lambda N +\lVert H_b \rVert+\lVert B \rVert$, where $N$ is the number of pixels. It will be large for high resolution images and  huge for good approximations of the $\MK$ cost (\emph{i.e.} $\lambda \gg1$). 
Such a scheme may thus involve a very slow explicit gradient ascent in the dual update \eqref{algo:primaldual}. %{time_step}
In such a case, we can resort to the alternative scheme proposed in the next subsection.%\vspace*{-0.1cm}

\subsection{Optimization using proximity operator of $\MK^*_{\lambda^*}$}
\label{sec:prox_sinkhorn_distance}

An alternative optimization of \eqref{eq:OT_segmentation_energy_reg} 
consists in using the proximity operator of $\MK_\lambda^*$.
Since we cannot compute the proximity operator of $\MK_\lambda^*$ in a closed form, 
we resort instead to a bidualization, as previously done in Section~\ref{sec:bidualization}.

Considering now the normalized function $\MK_\lambda (\alpha)$ using entropy normalization~\eqref{eq:normalized_entropy} on set $\S$,
we thus have $\MK_\lambda^* (\beta) = \frac{N}{\lambda} \langle Q_\lambda(\beta), \U\rangle =  g_\lambda^*(L \beta)$.
\begin{proposition}\label{prop:prox_bidual}
	The proximity operators of 
	$
	g_\lambda^*(q) = \frac{N}{\lambda} \dotp{e^{\lambda (q-c) - \U}}{\U}
	$
	is
	\begin{equation}\label{eq:prox_dual}
	\text{prox}_{\tau g_\lambda^*}(p) = p - \frac{1}{\lambda} W\left(  \lambda \tau N e^{\lambda (p-c) - \U}  \right).
	\end{equation}
where $W$ is the \emph{Lambert function}, such that $w=W(z)$ is solution of $we^w = z$. The solution is unique as $z=\lambda \tau N e^{\lambda (p-c) - \U}\geq 0$.
%\vspace*{-1mm}
\end{proposition}
\begin{proof}
See appendix \ref{proof:prox_bidual}.
%\vspace*{-1mm}
\end{proof}

\begin{remark}
Note that the Lambert function can be evaluated very fast.
%\vspace*{-2mm}
\end{remark}

%%%%%%%%%%%%%%%%%%%%%%%%%%%%%%%%%%%%%%%%%%%%%%%%%%%

\section{Experiments}

\paragraph{Experimental setting}

\begin{figure}[phtb]
\centering{
\begin{tabular}{ccc}
\includegraphics[width=0.32\textwidth]{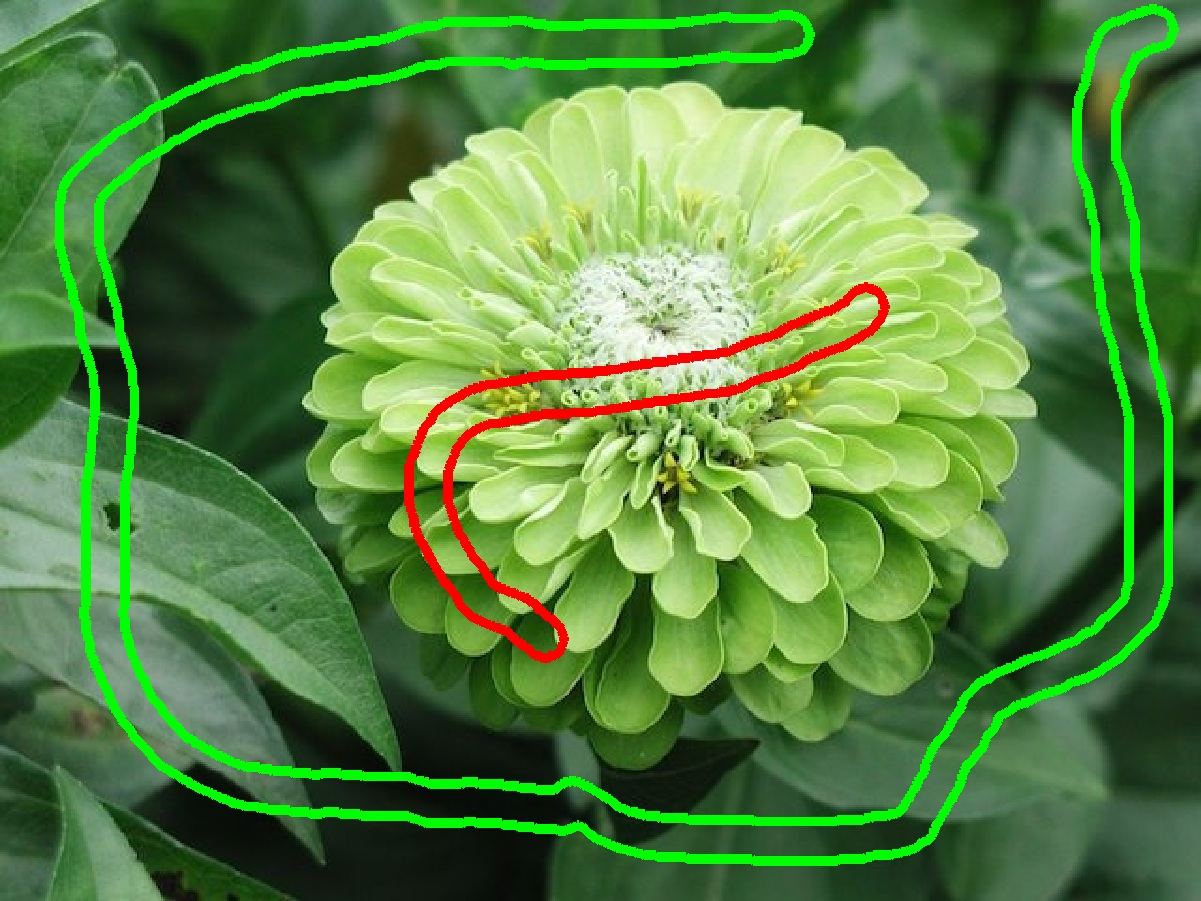} &
\includegraphics[width=0.32\textwidth]{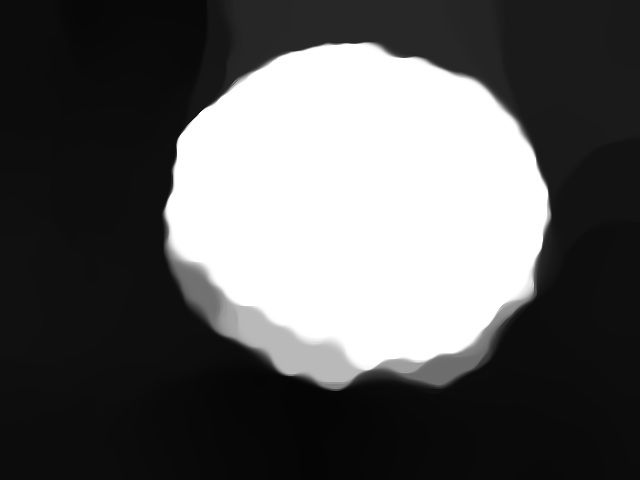} &
\includegraphics[width=0.32\textwidth]{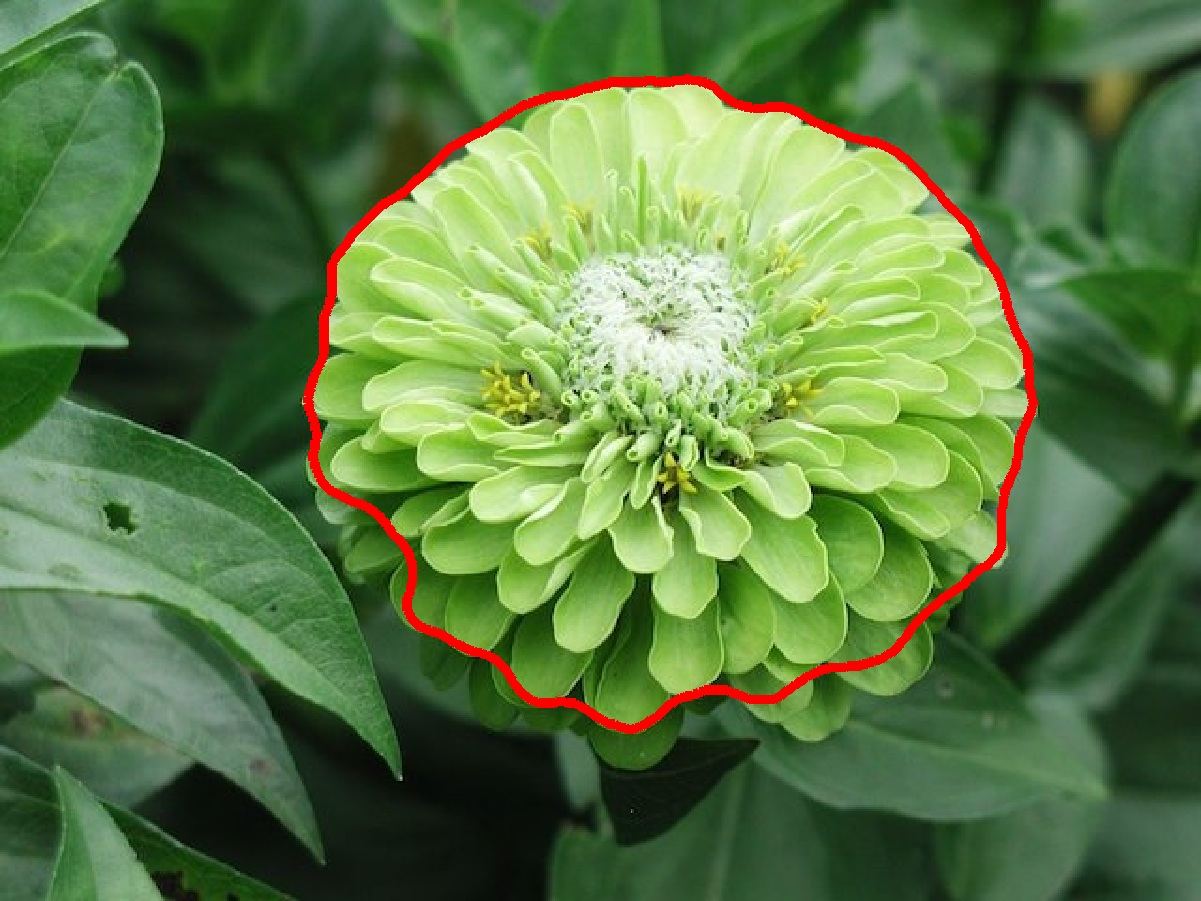}
\\
Input & $u$  & $t=0.5$ 
\\
\includegraphics[width=0.32\textwidth]{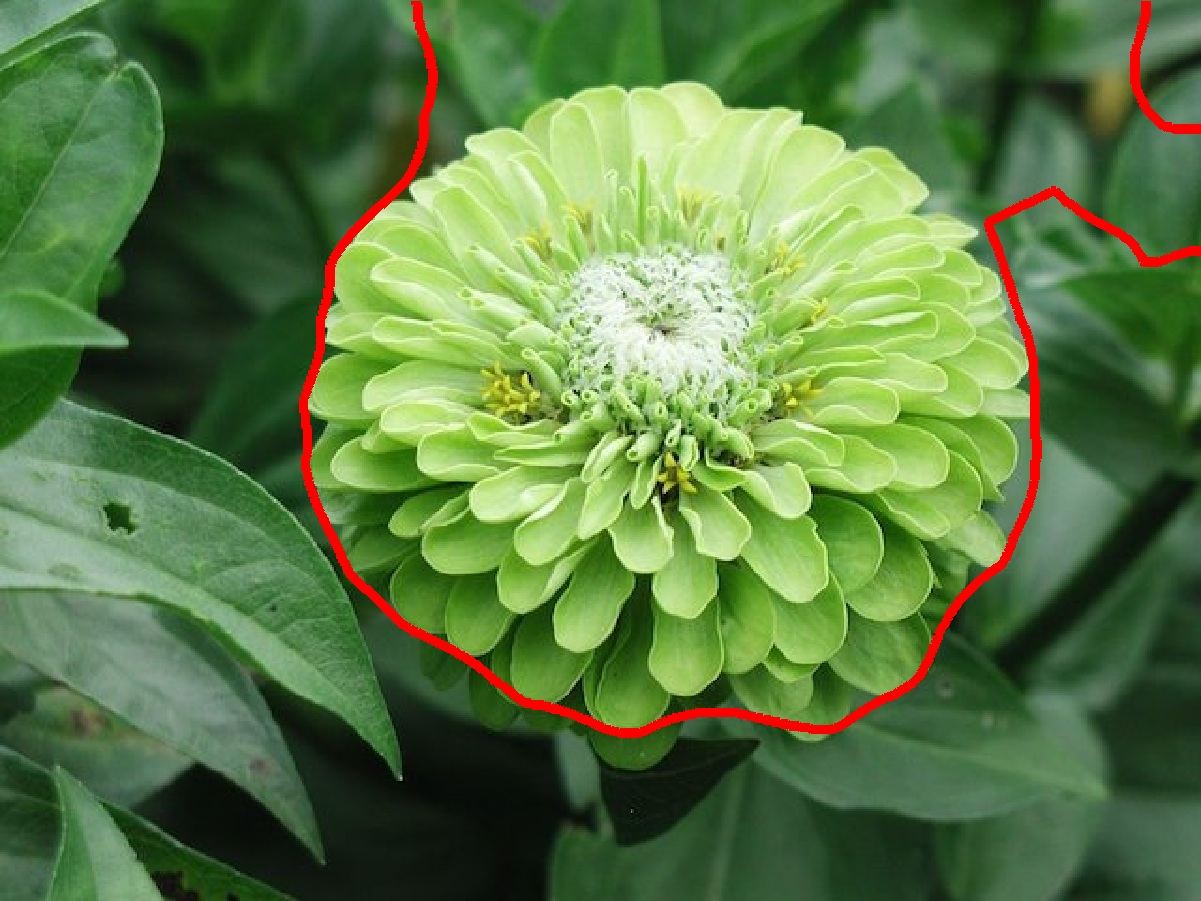} &
\includegraphics[width=0.32\textwidth]{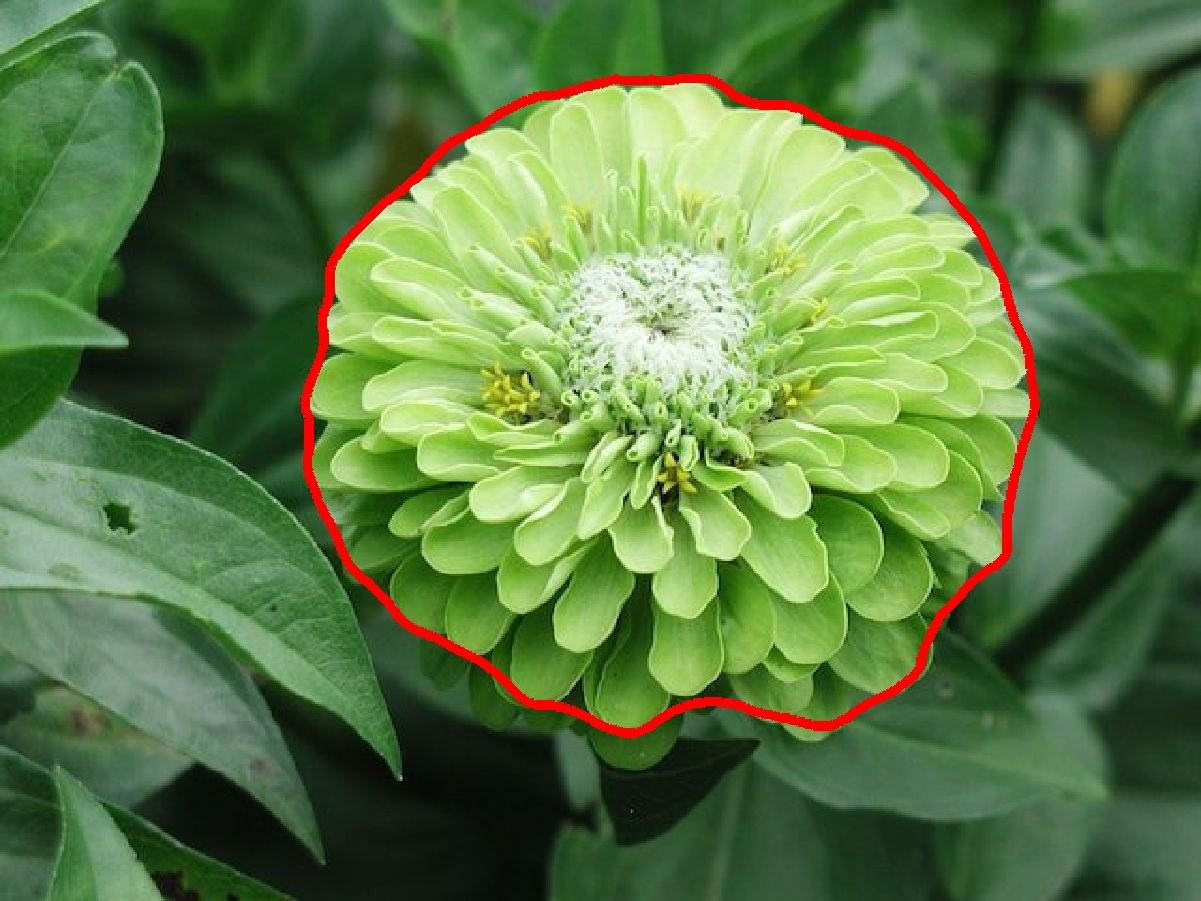} & 
\includegraphics[width=0.32\textwidth]{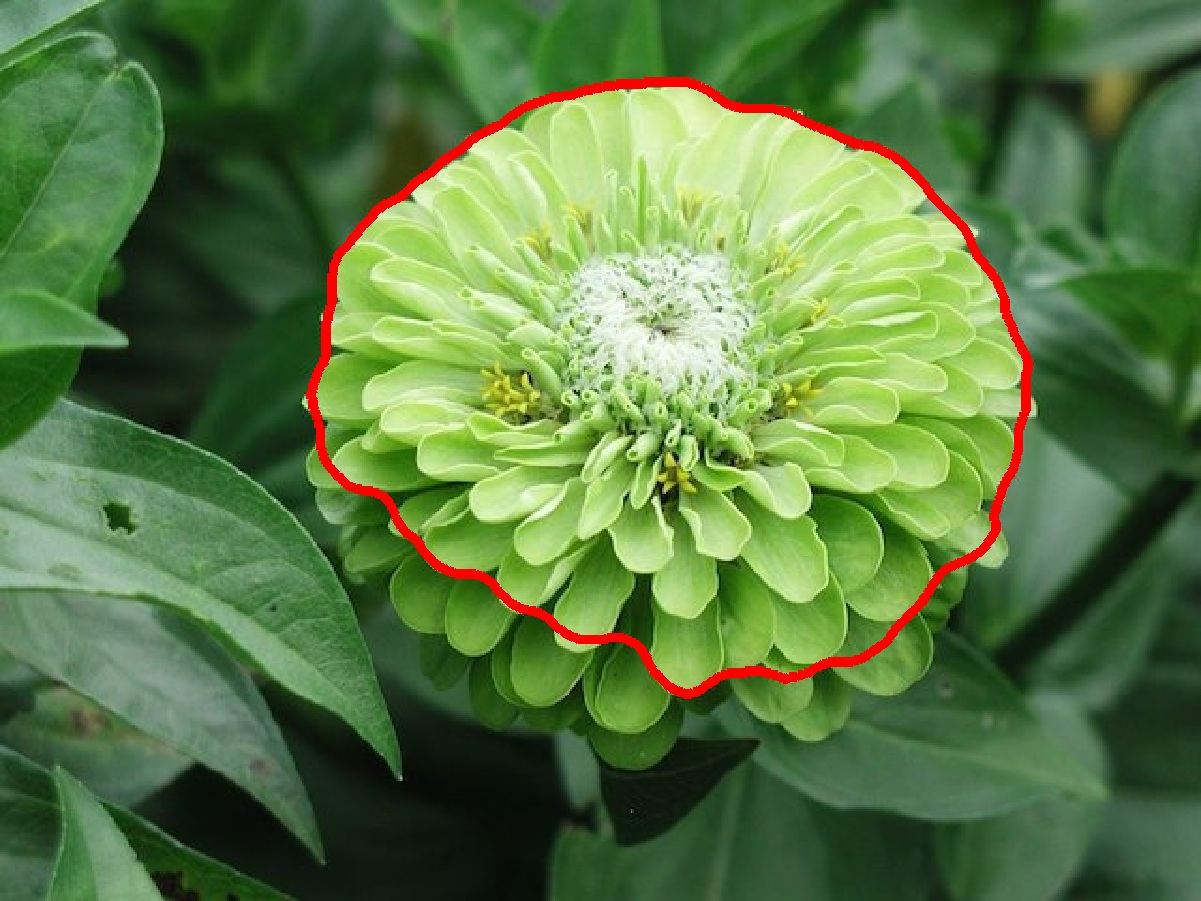} 
\\
$t=0.1$ & $t=0.2$ & $t=0.9$
\end{tabular}
}
\caption{Illustration of the image segmentation method and the influence of the final threshold parameter $t$ on the segmentation result (here using non regularized optimal transport, \emph{i.e.} $\lambda = \infty$).
The user defines scribbles which indicates the object to be segmented (here in red) and the background to be discarded (in green). 
The output image $u$ is a regularized weight map that gives the probability of a pixel to belong to the object.
This probability map $u$ is finally thresholded with a parameter $t$ to segment the input image into a region $R_t(u)$, which contour is displayed in red.
In the rest of the paper, $t=0.5$ is always used, but other strategies may be defined, such as selecting the threshold value that minimizes the non-relaxed energy \eqref{eq:segmentation_energy}.
}
\label{fig:thres}
\end{figure}

\begin{figure}[phtb]
\centering
	\begin{tabular}{cccc}
	\includegraphics[width=0.32\textwidth]{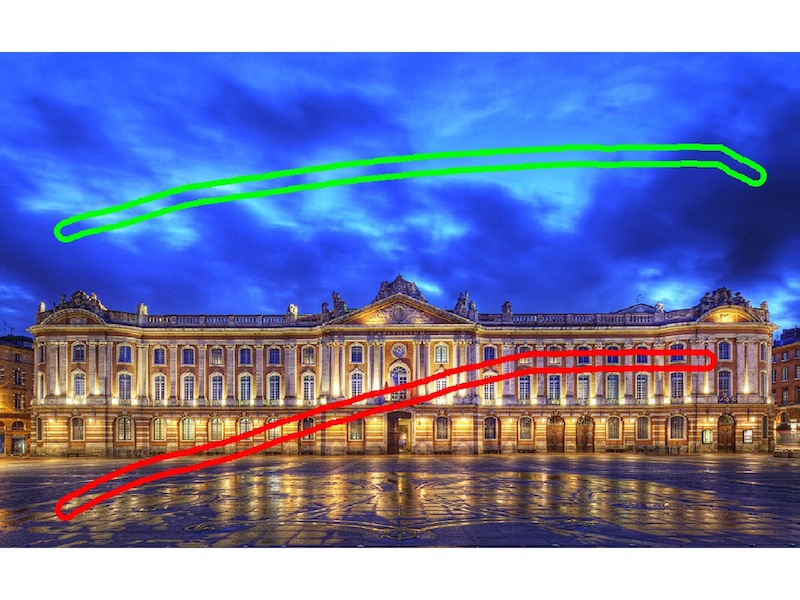} &
	\includegraphics[width=0.32\textwidth]{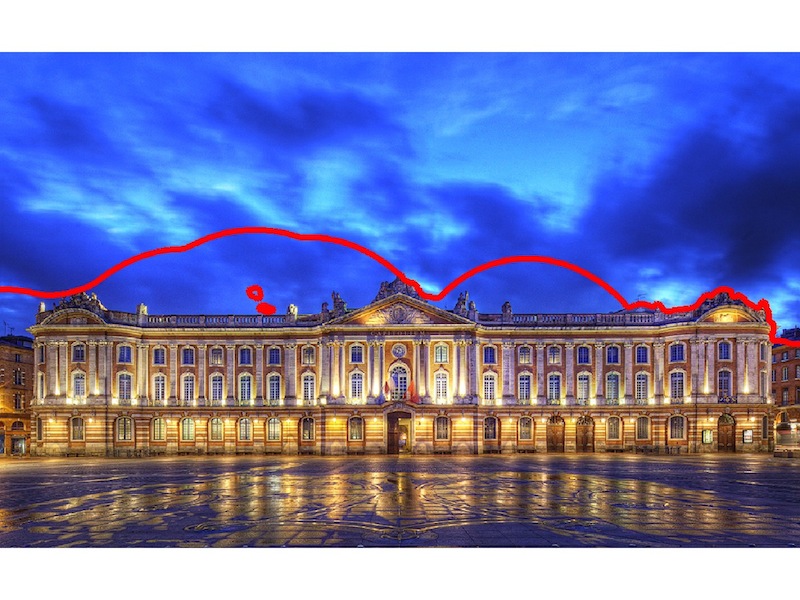} &
	\includegraphics[width=0.32\textwidth]{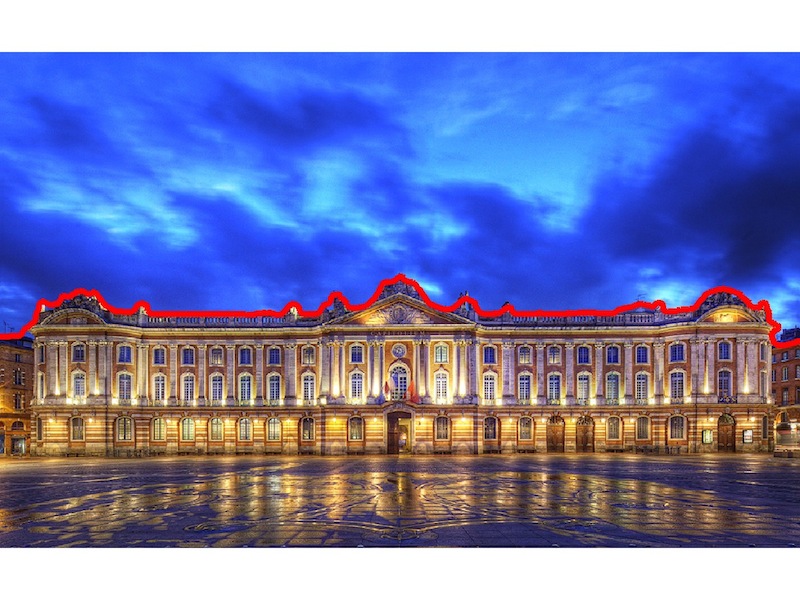}
	\\
	Input & $L_1$ & OT ($\lambda = \infty$)
	\end{tabular}

\caption{Robustness of OT with respect to $L_1$: the blue colors that are not in the reference histograms are considered as background with OT distance and  as foreground with the $L_1$ model, since no color transport is taken into account.
}
\label{fig:comp_l1}
\end{figure}

\begin{figure}[!p]
\centering
\begin{tabular}{cccc}
\includegraphics[width=0.24\textwidth]{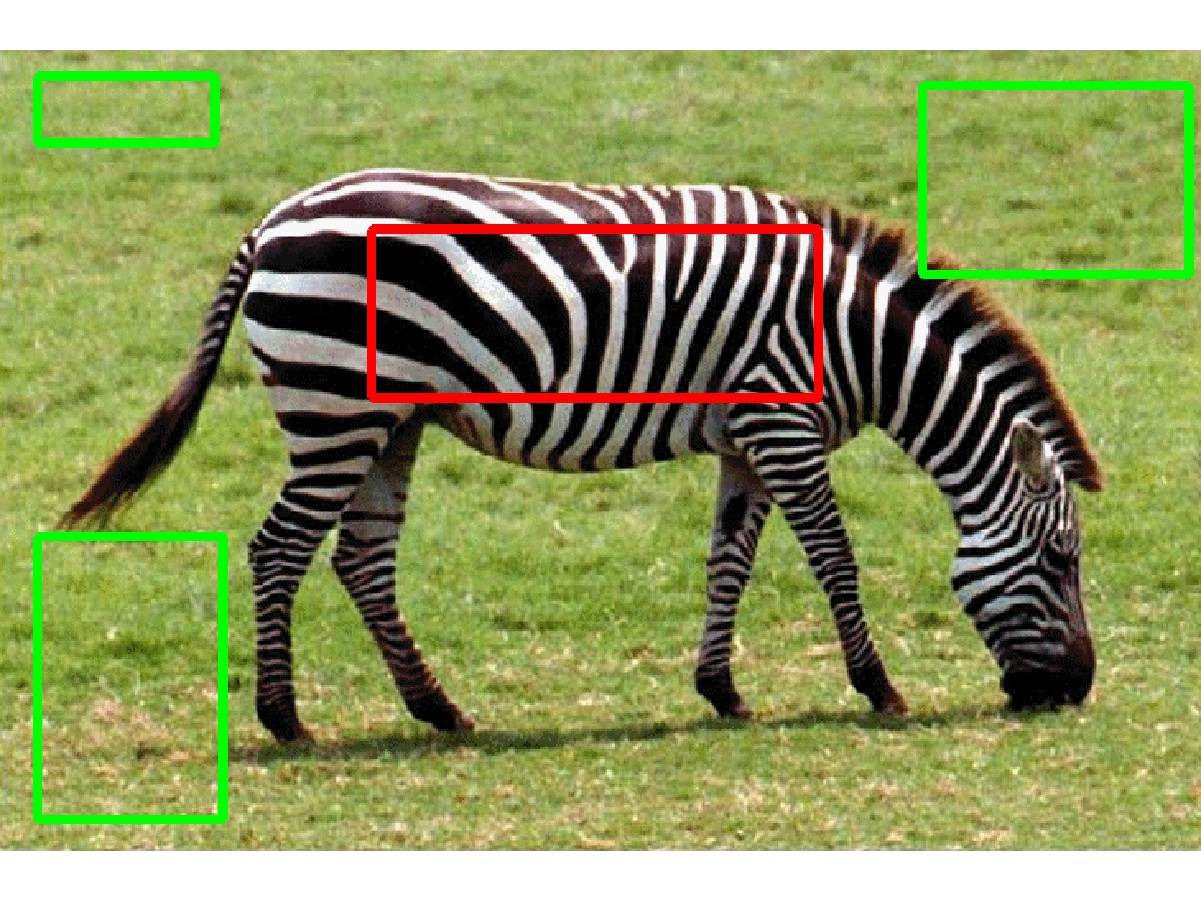} &
\includegraphics[width=0.24\textwidth]{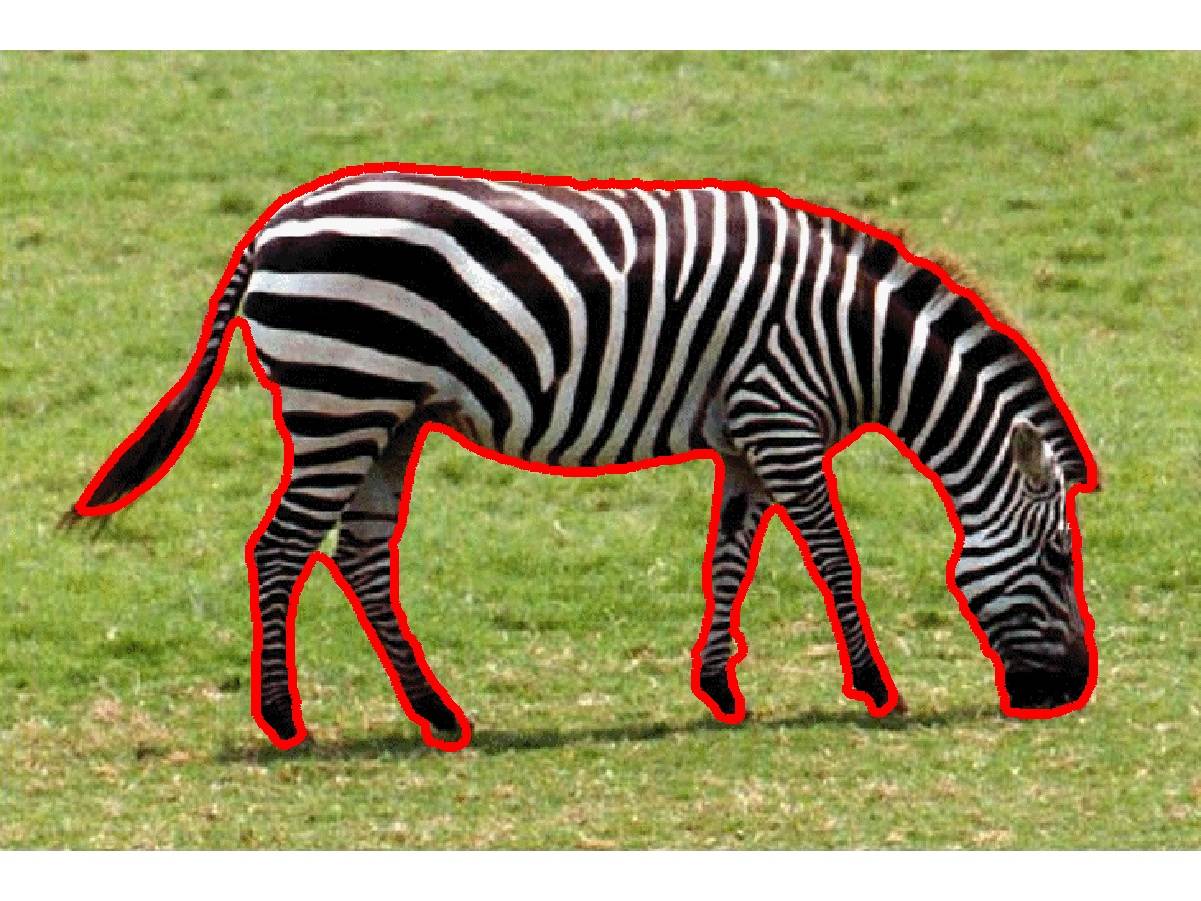} &
\includegraphics[width=0.24\textwidth]{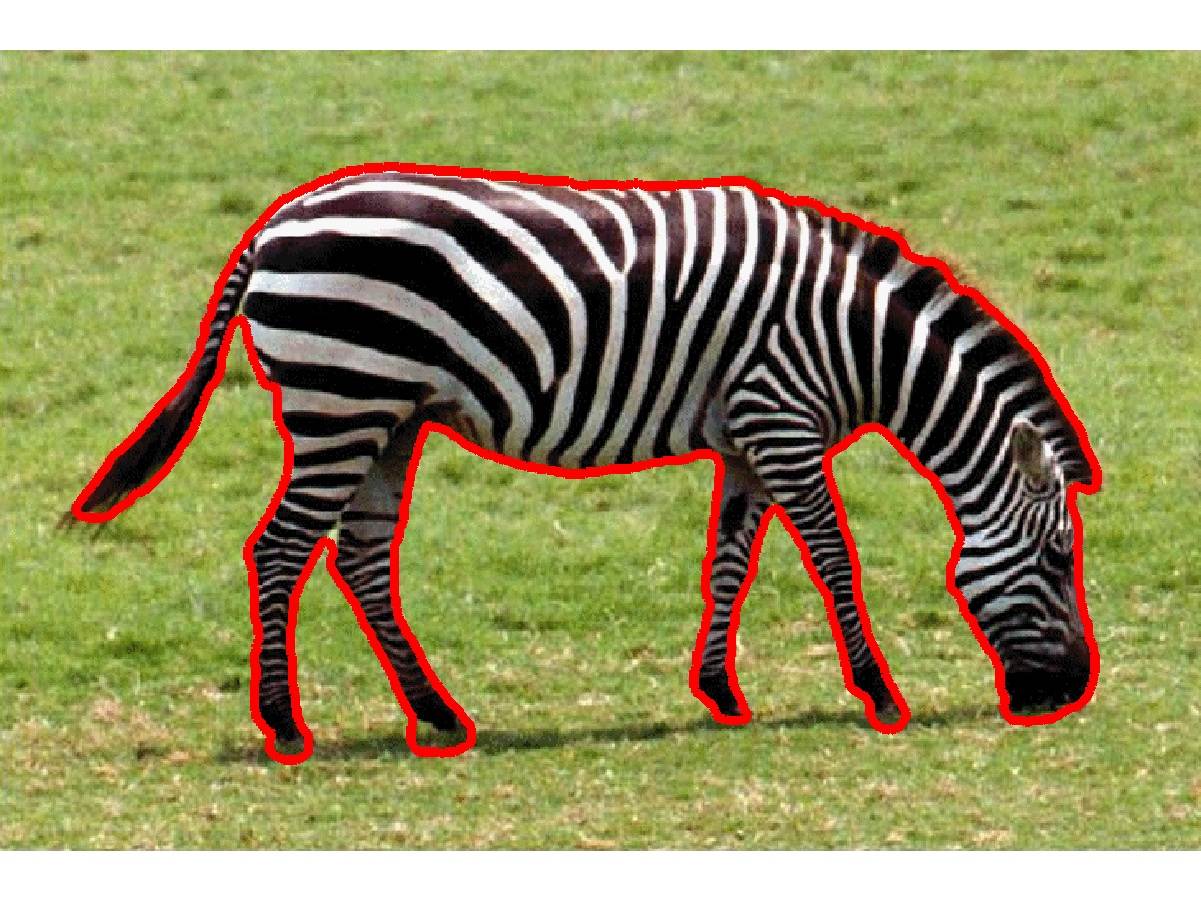} &
\includegraphics[width=0.24\textwidth]{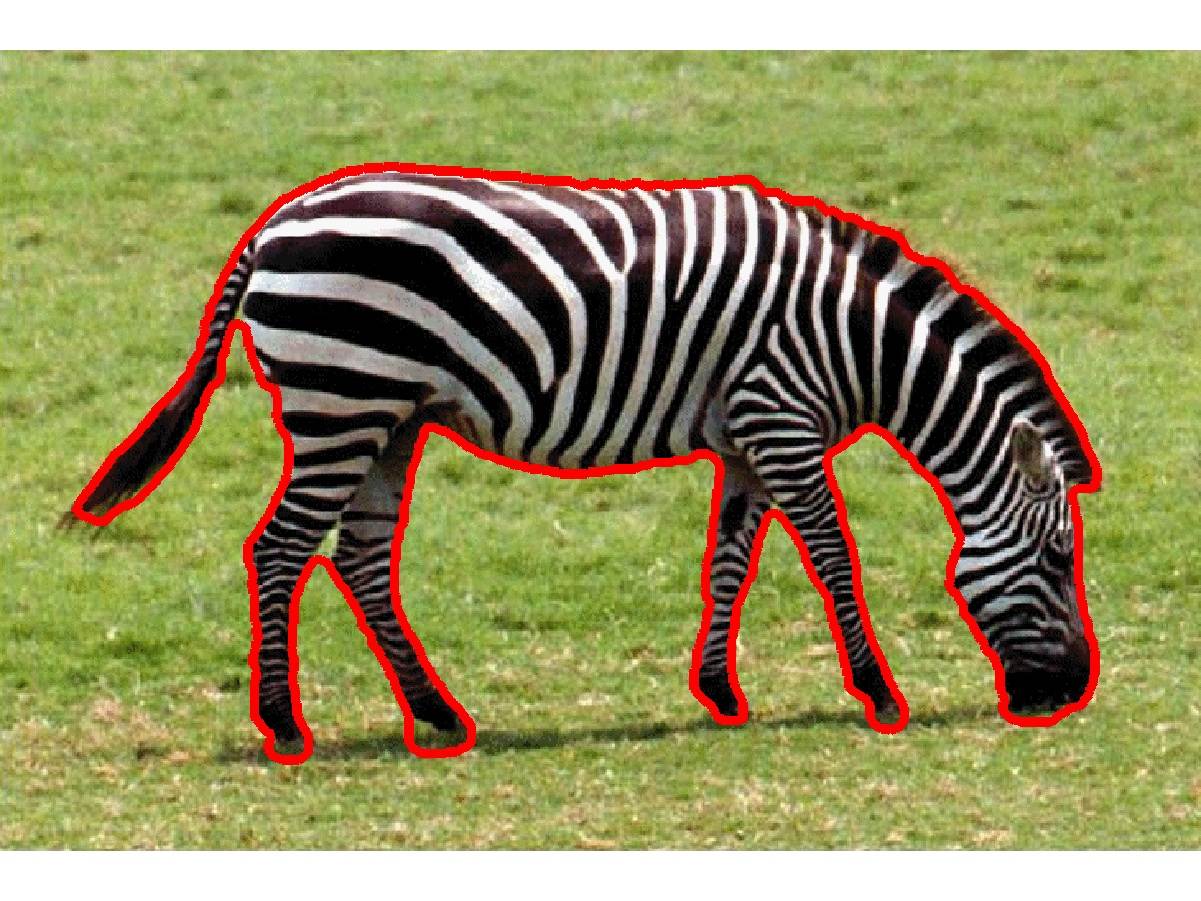} 
\\
\includegraphics[width=0.24\textwidth]{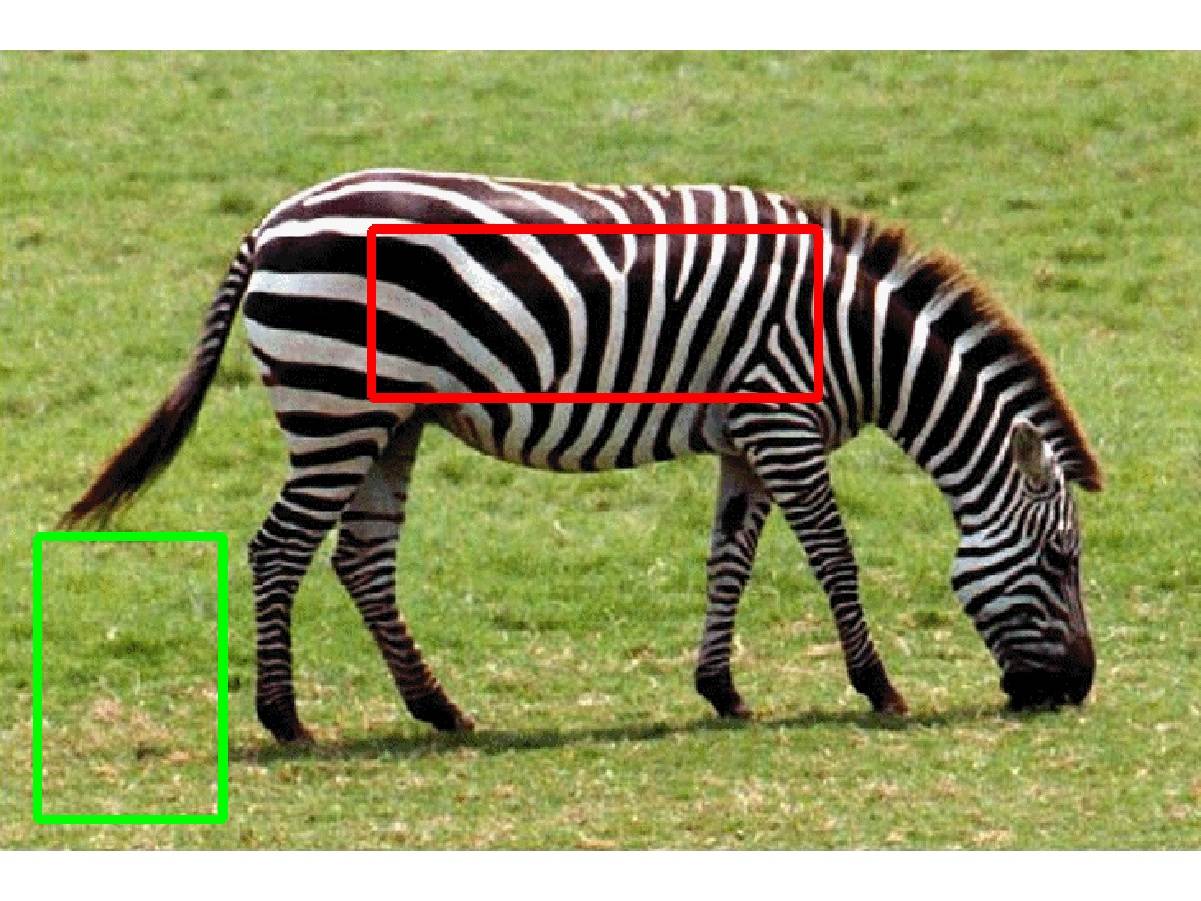} &
\includegraphics[width=0.24\textwidth]{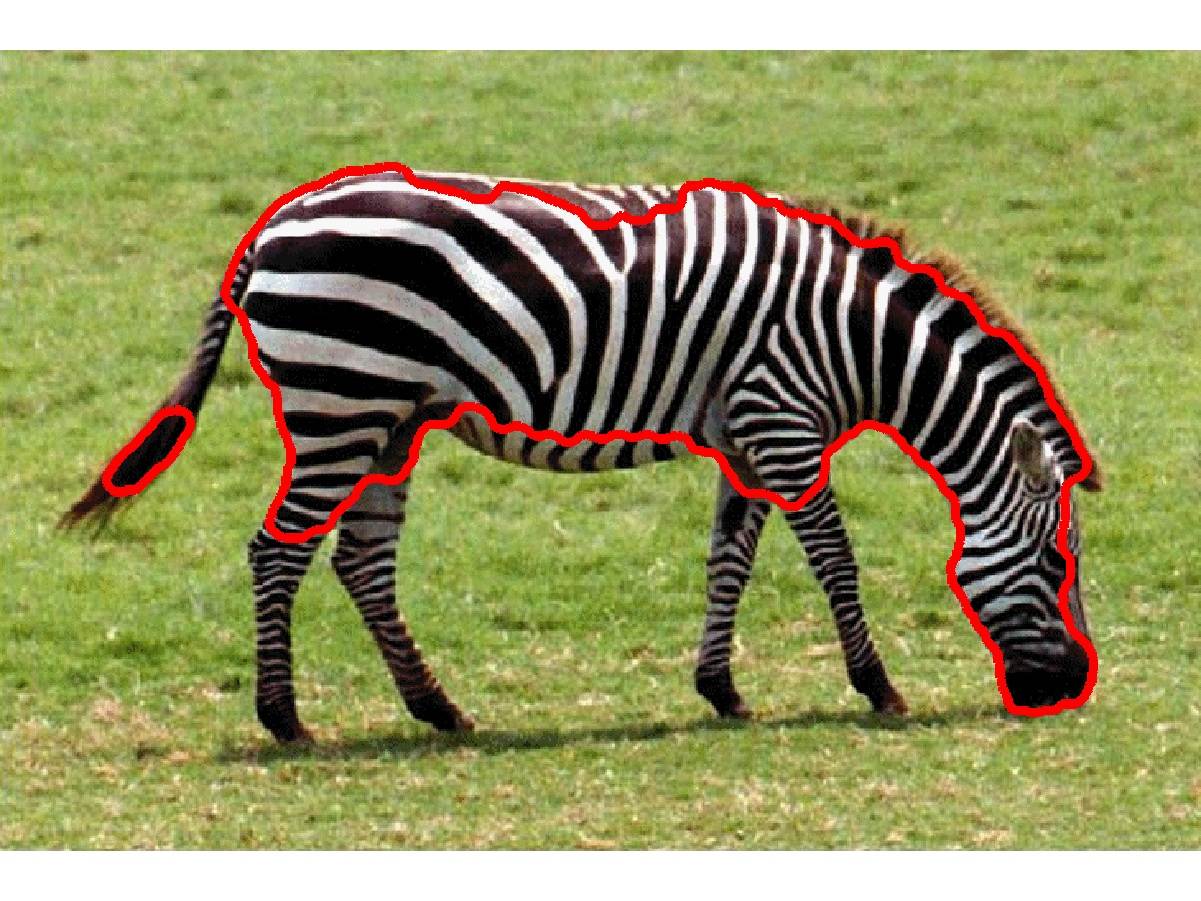} &
\includegraphics[width=0.24\textwidth]{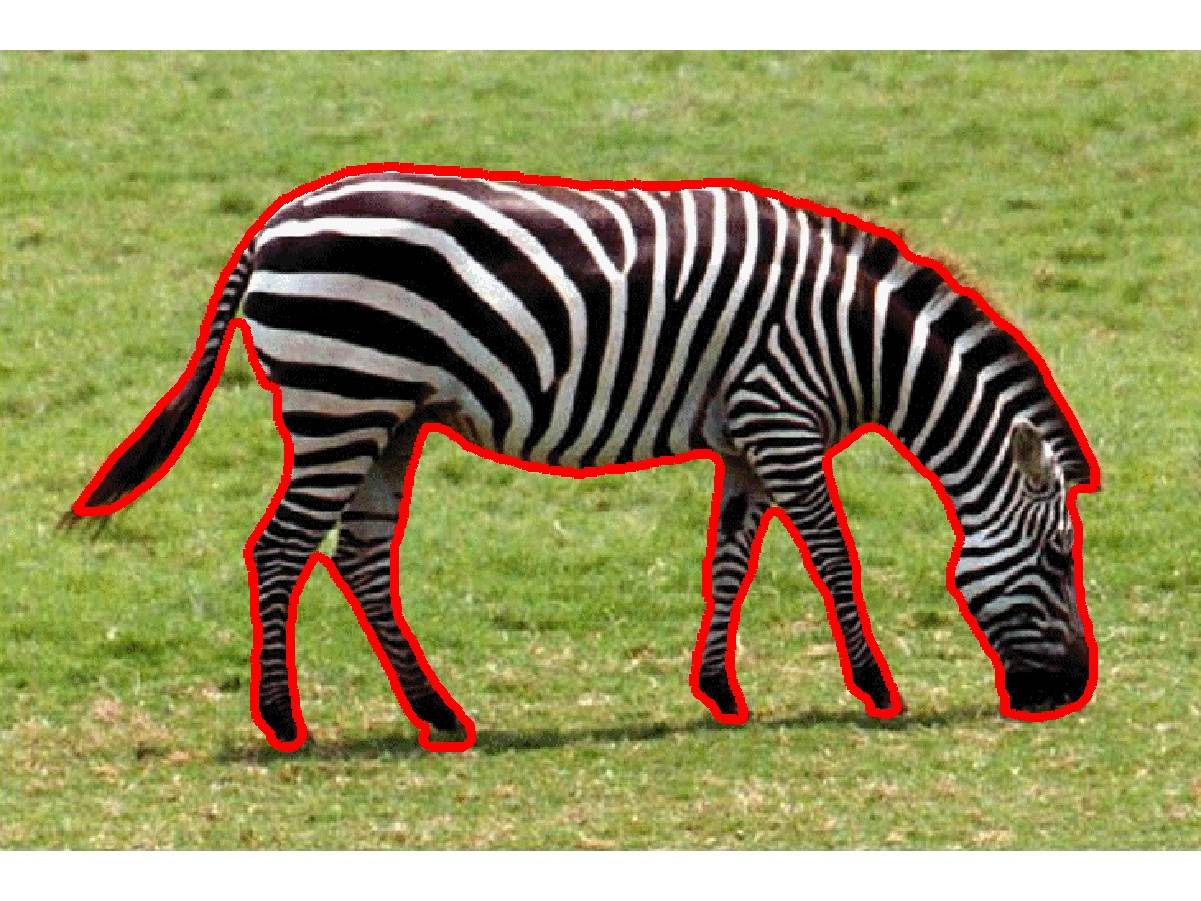} &
\includegraphics[width=0.24\textwidth]{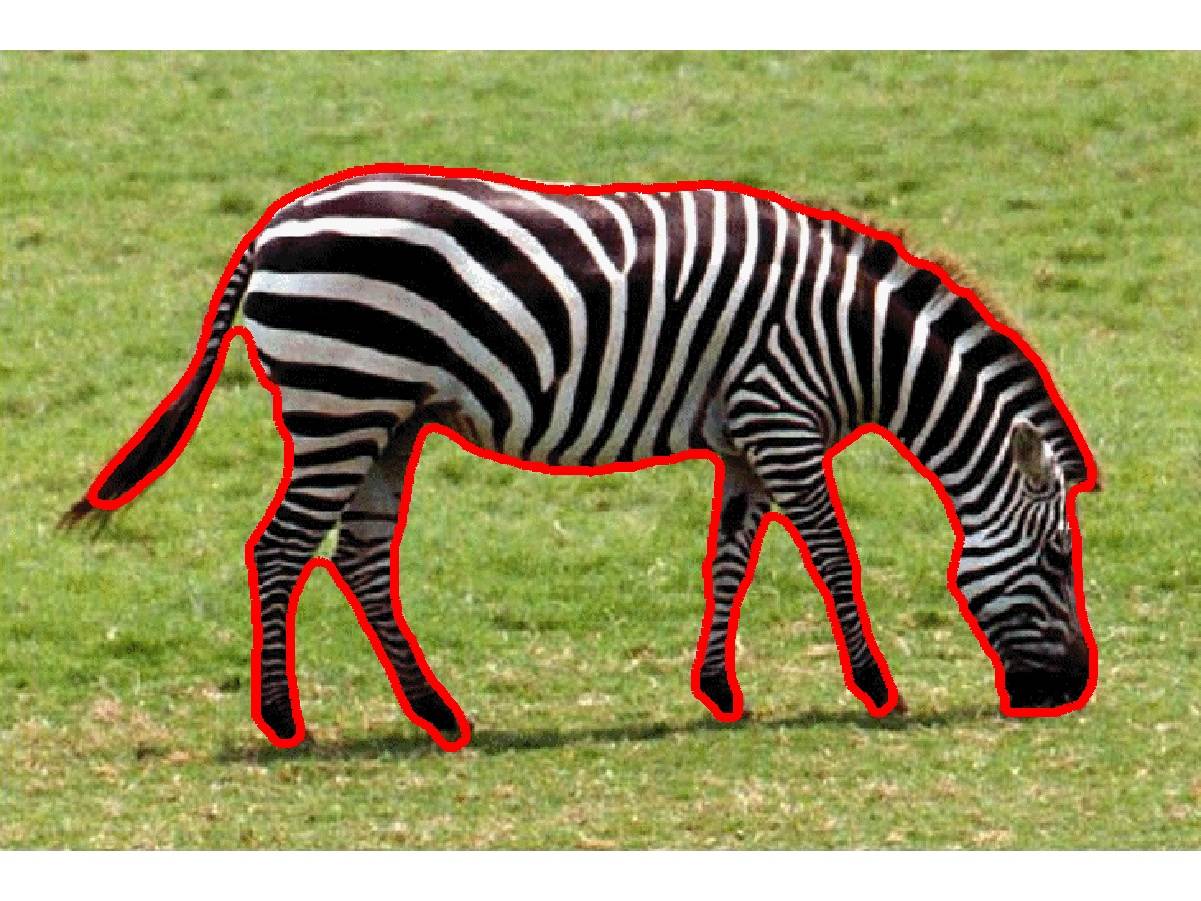} \\
Inputs & $\lambda = L_1$ & $\lambda = \infty$ & $\lambda = 1000$
\end{tabular}
\caption{Comparison of the segmentation results obtained from the proposed segmentation models (using $\MK_\lambda$ distances) together with the $L_1$ distance used in \cite{papa_aujol}. The same regularization parameter $\rho$ is used for every segmentations. Note that the optimal transport similarity measure is a more robust statistical metric between histograms than $L_1$.}
\label{fig:zeb}
\end{figure}

\begin{figure}[!p]
\centering{
	\begin{tabular}{cccc}
	\includegraphics[width=0.24\textwidth]{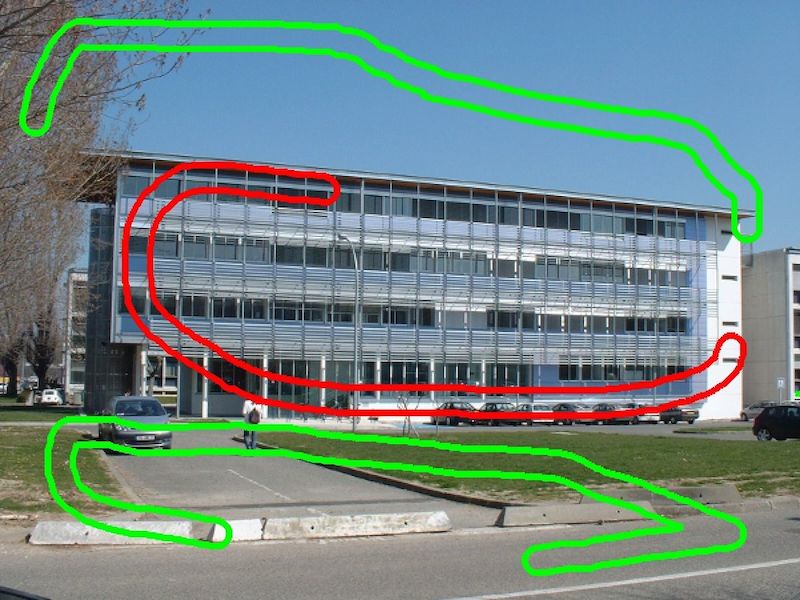} &
	\includegraphics[width=0.24\textwidth]{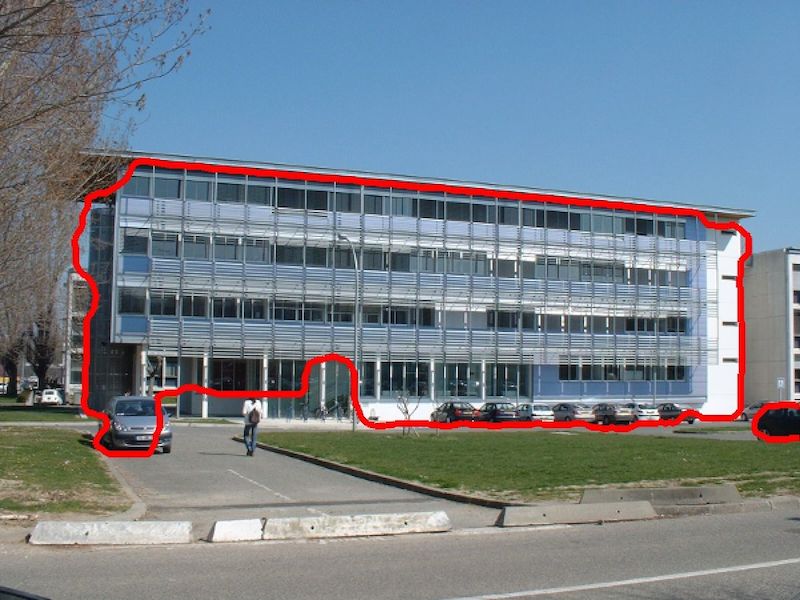} &
	\includegraphics[width=0.24\textwidth]{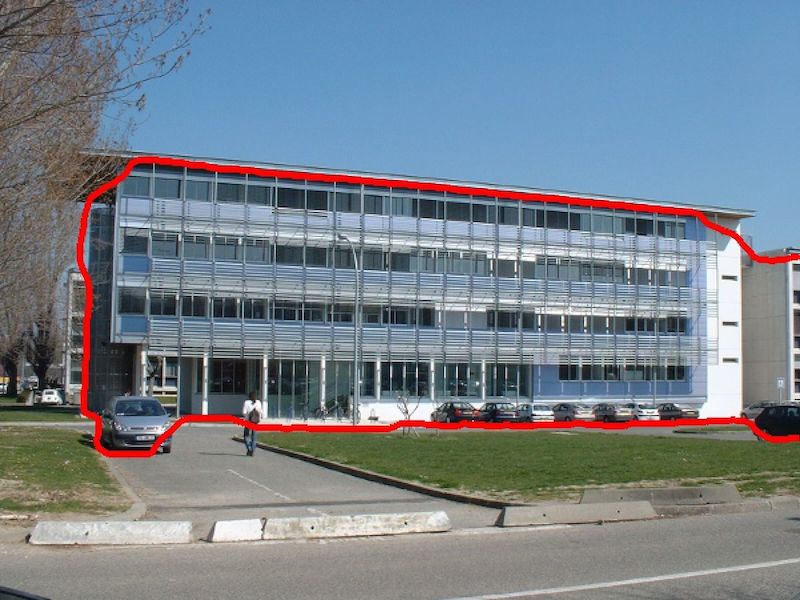} &
	\includegraphics[width=0.24\textwidth]{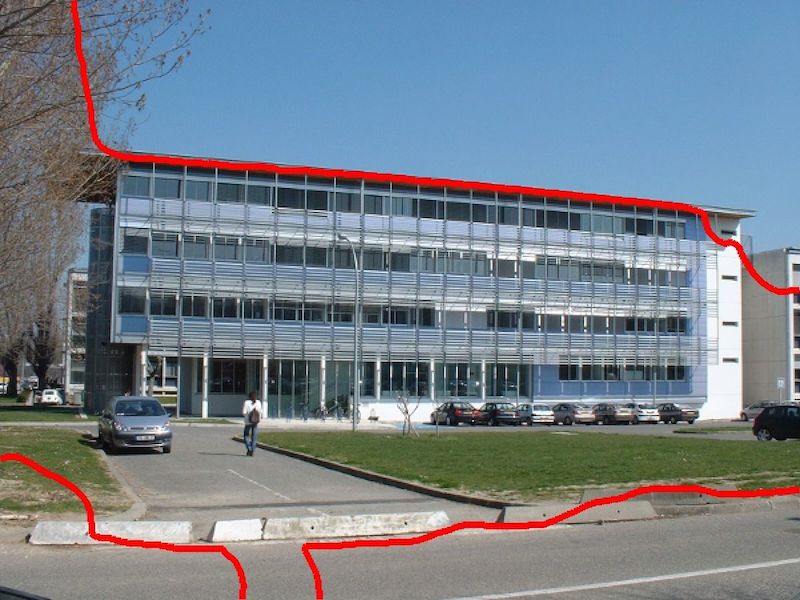} 
	\\
%	Input & $\lambda = \infty$ & $\lambda = 1000$ & $\lambda = 100$\\
	\includegraphics[width=0.24\textwidth]{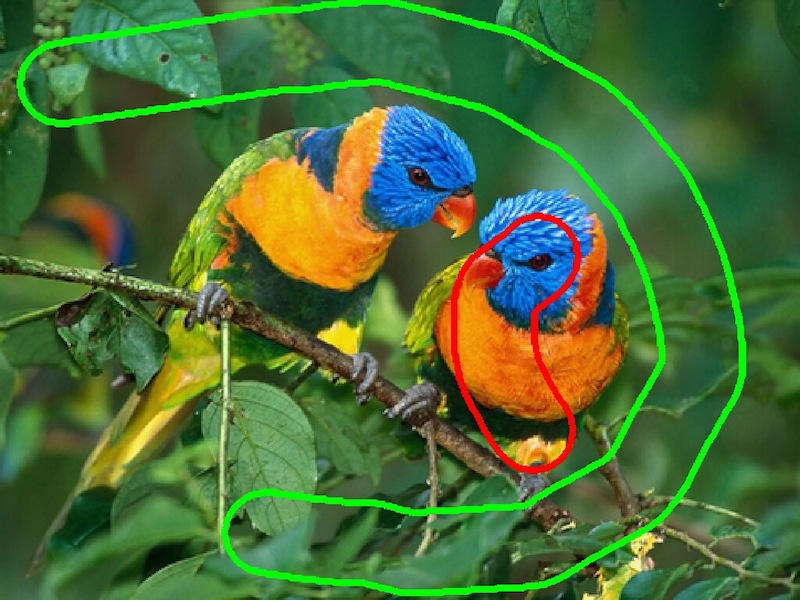} &
	\includegraphics[width=0.24\textwidth]{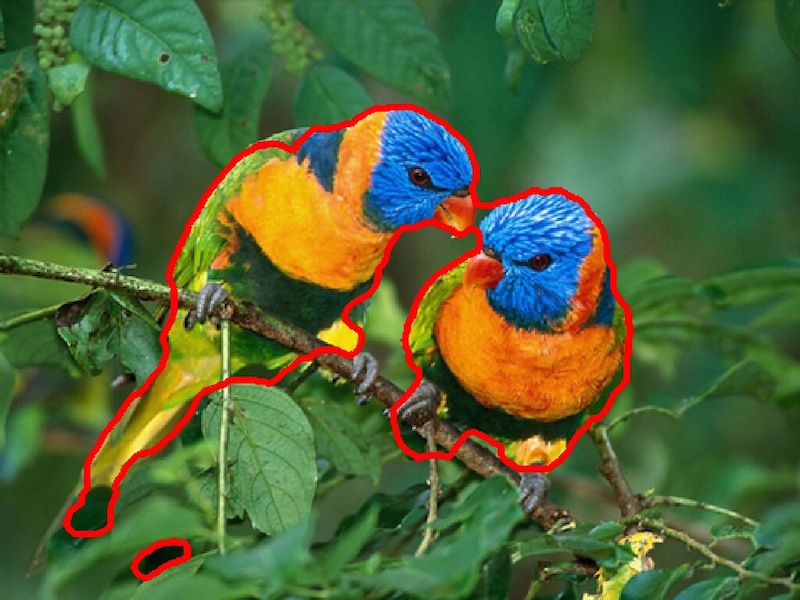} &
	\includegraphics[width=0.24\textwidth]{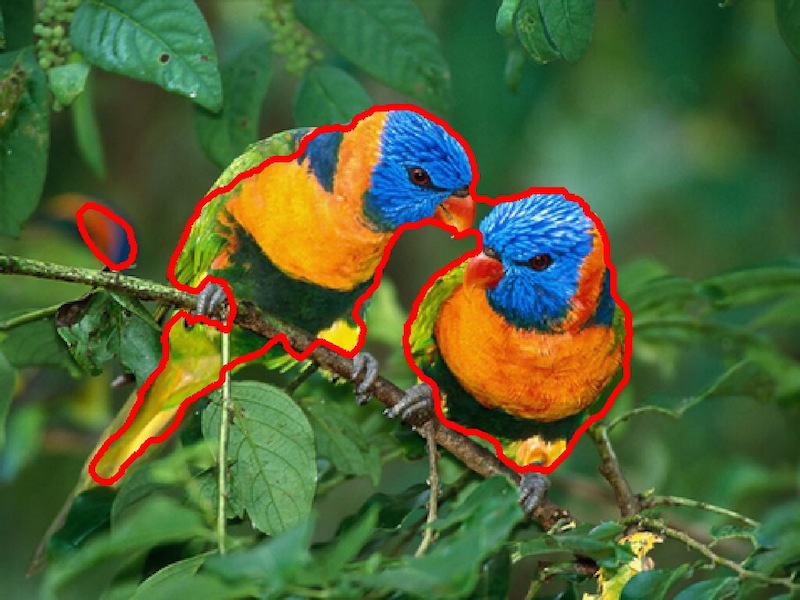} &
	\includegraphics[width=0.24\textwidth]{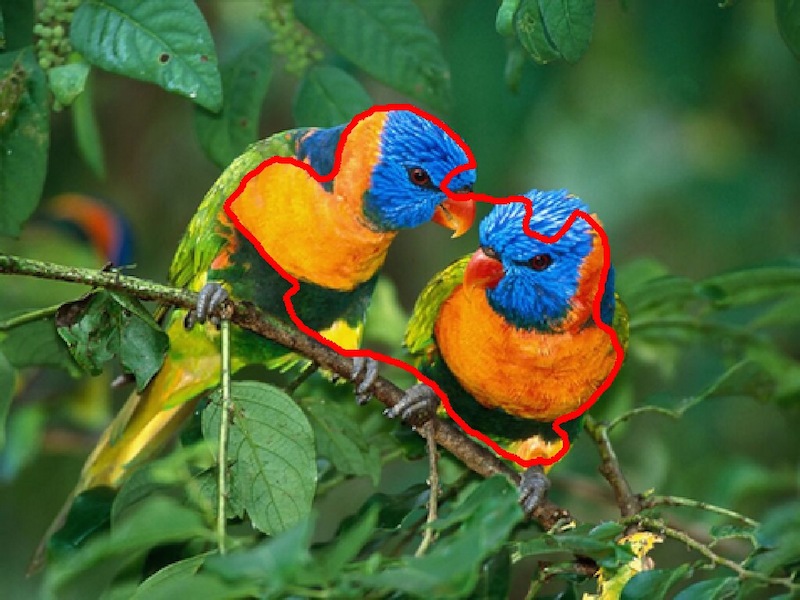} 
	\\
	\includegraphics[width=0.24\textwidth]{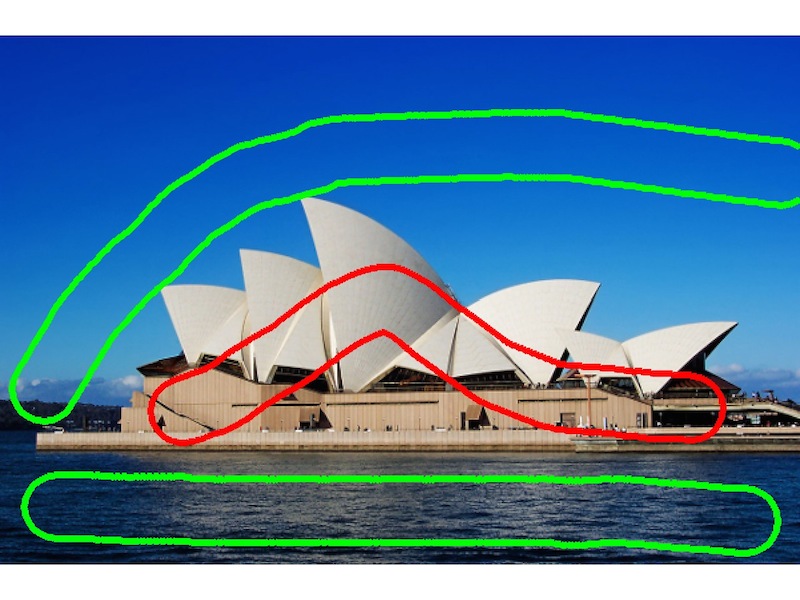} &
	\includegraphics[width=0.24\textwidth]{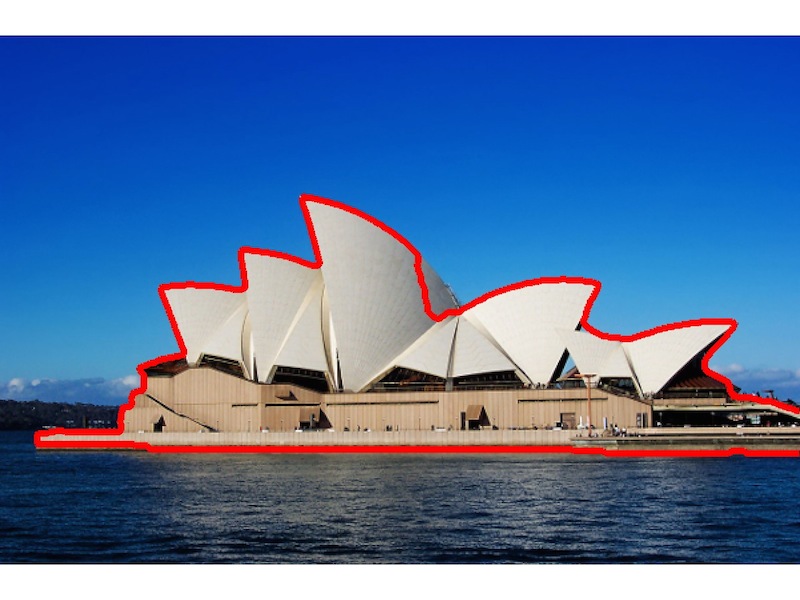} &
	\includegraphics[width=0.24\textwidth]{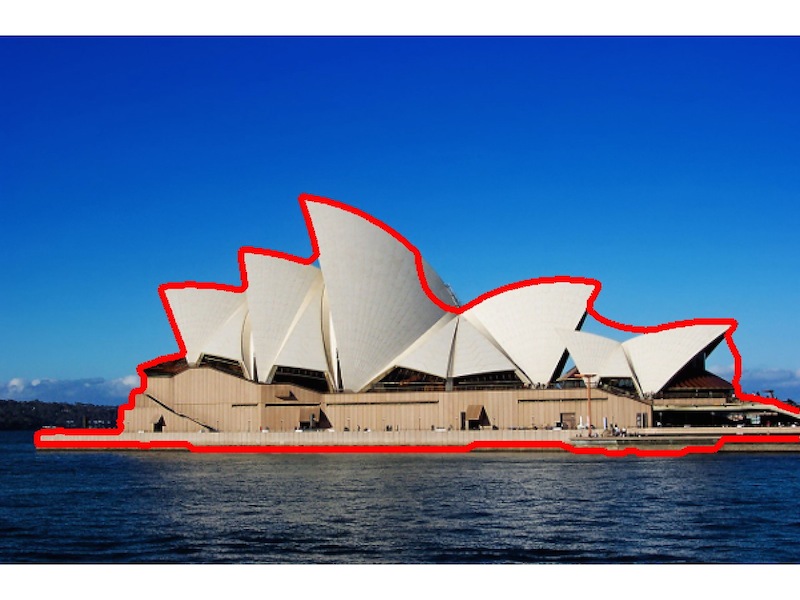} &
	\includegraphics[width=0.24\textwidth]{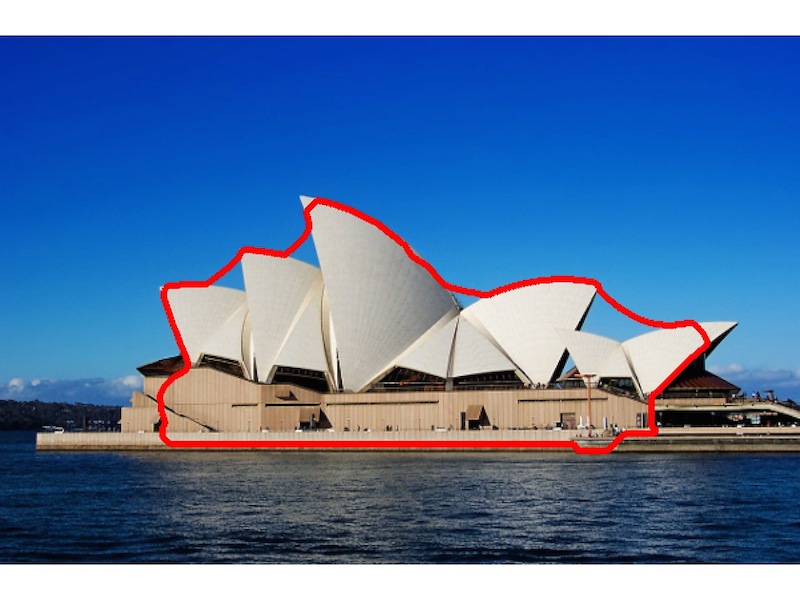} 
	\\
	Input & $\lambda = \infty$ & $\lambda = 100$ & $\lambda = 10$\\	
	\end{tabular}
}

\caption{Comparison of segmentations obtained from the proposed models. The input areas are used to compute the reference color distributions $a$ and $b$. The non-regularized model corresponds to $\lambda=+\infty$, increasing regularization effects are then shown.
}
\label{fig:comp_lambda}
\end{figure}

\begin{figure}[p]
\centering
\begin{tabular}{cc}
\includegraphics[width=0.36\textwidth]{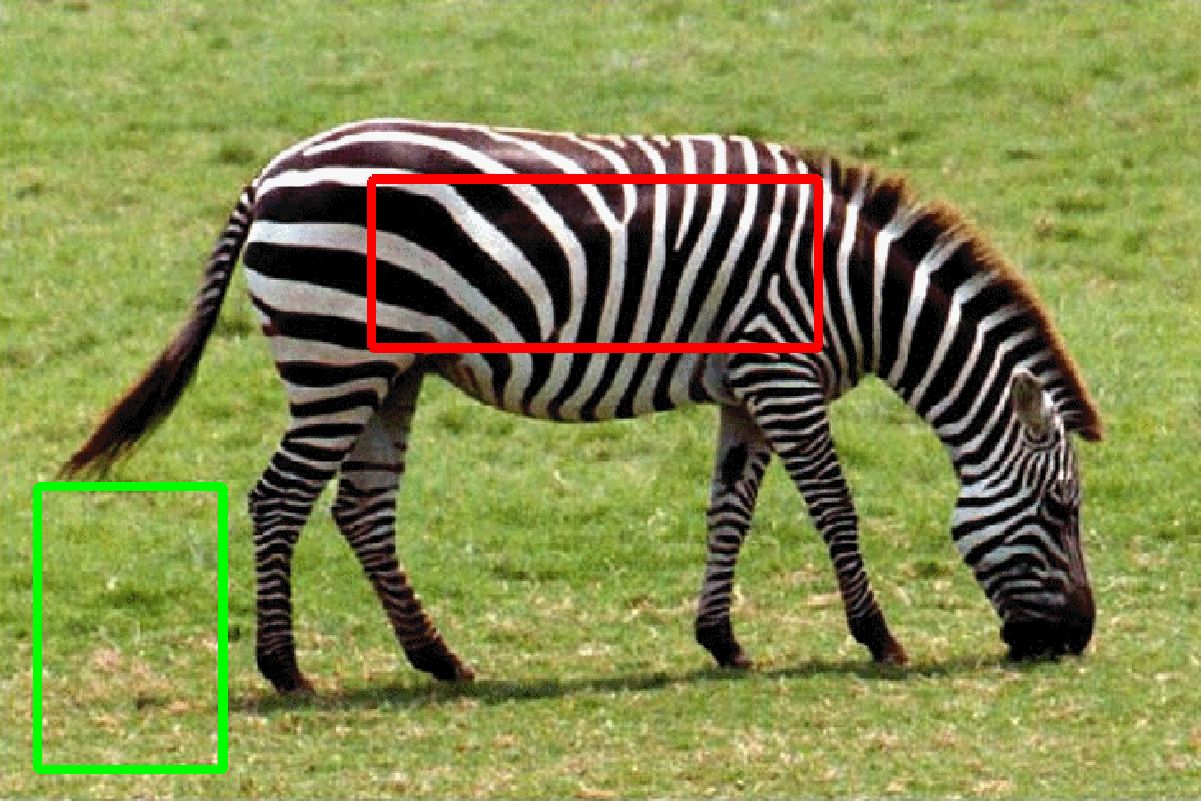} &
\includegraphics[width=0.36\textwidth]{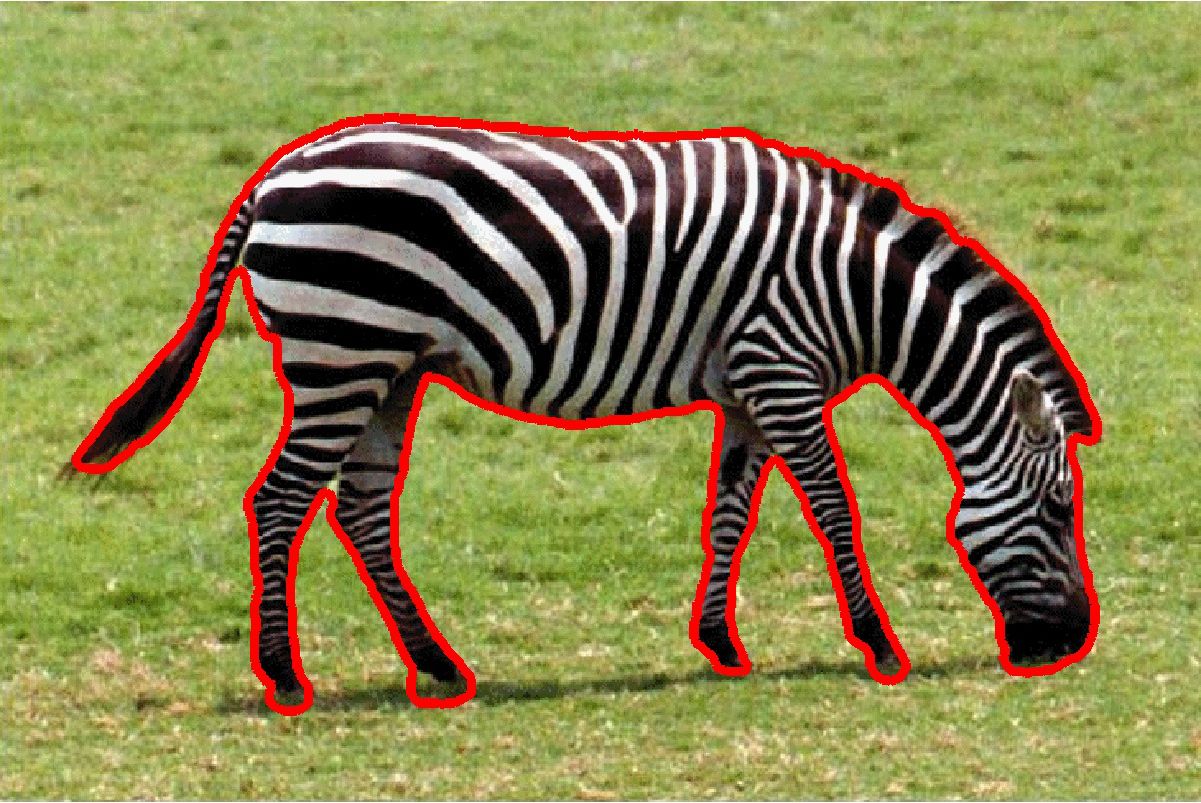}
\\
Input histogram & Resulting segmentation
\\[2mm]
\includegraphics[width=0.34\textwidth]{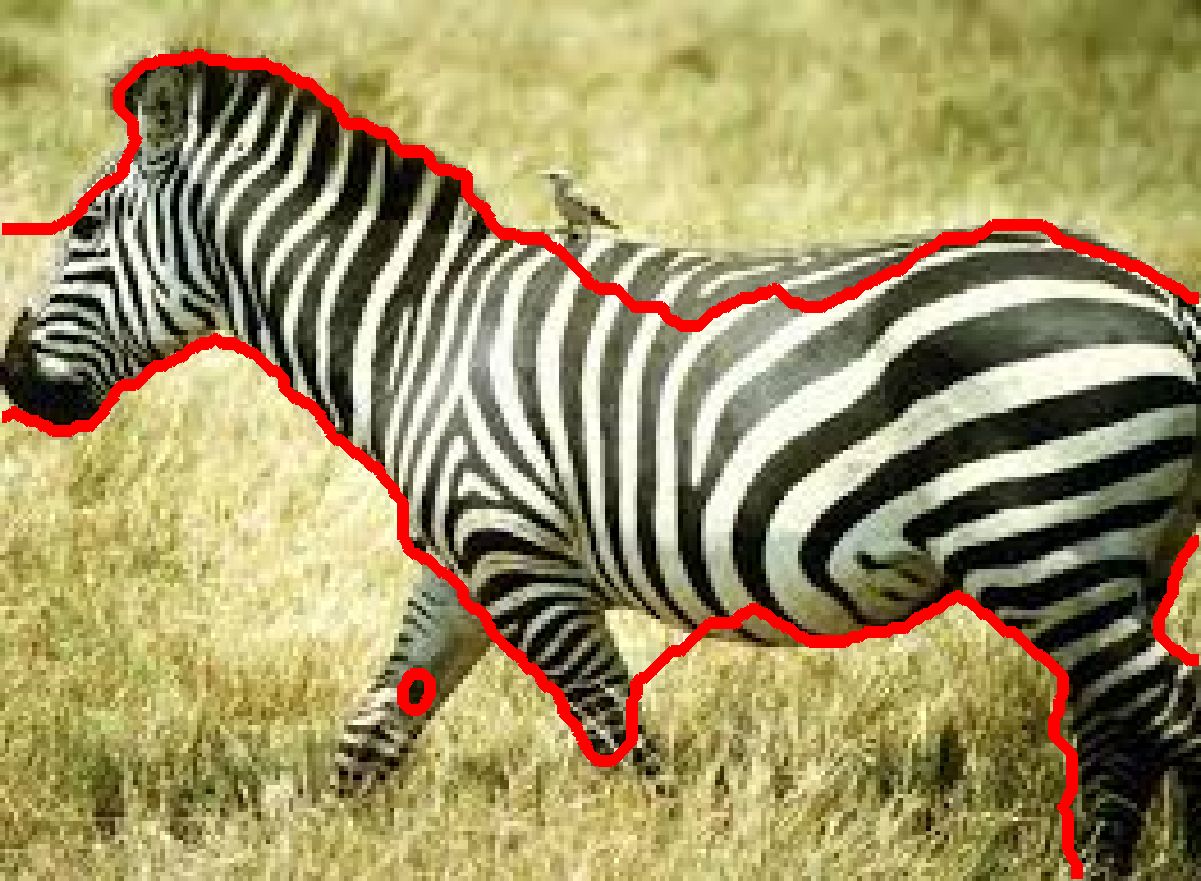} &
\includegraphics[width=0.4\textwidth]{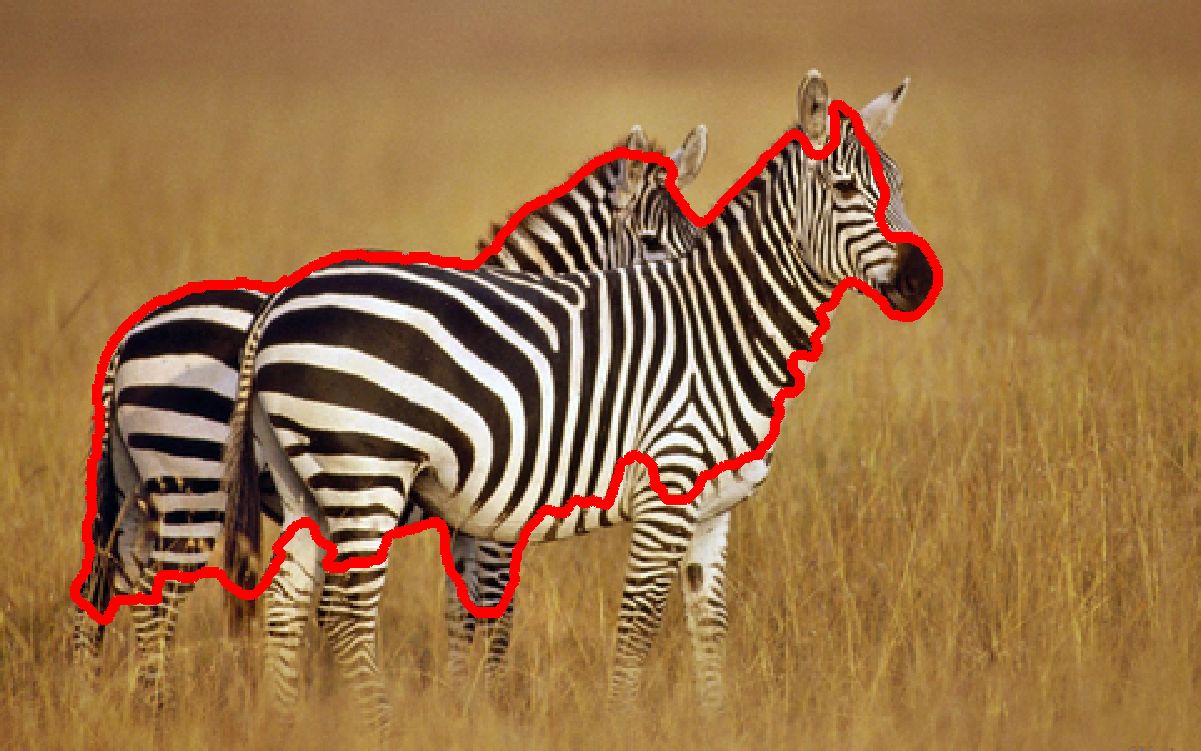}
\\
Segmentation on a different image & with the same exemplar histogram
\end{tabular}
\caption{Illustration of the interest of optimal transport for the comparison of histograms. Its robustness makes it possible to use prior histograms from different images (here histograms are taken from image $1$ and used to segment images $2$ and $3$), even with a different clustering of feature space. Note that it is not possible with bin-to-bin metric, which requires the same clustering.}
\label{fig:zebras}
\end{figure}

\begin{figure}[p]
\centering
	\begin{tabular}{cccc}
	\includegraphics[height=2.5cm]{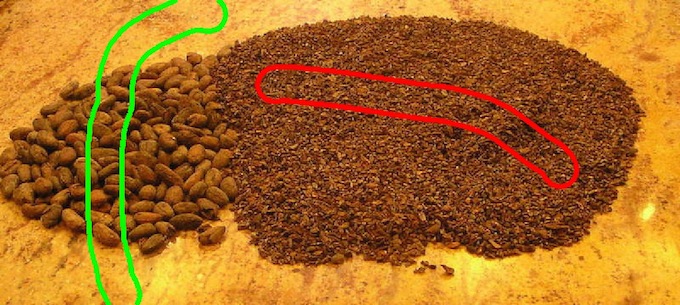} & 
	\includegraphics[height=2.5cm]{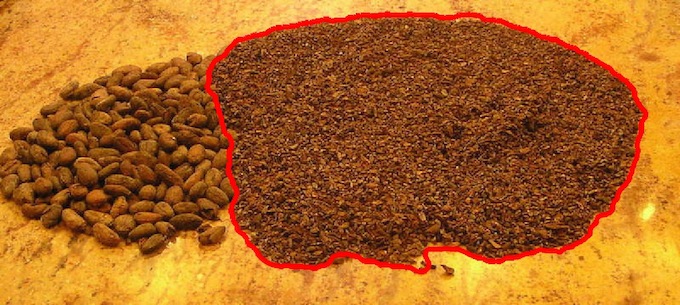} 
	\\[3mm]
	\includegraphics[height=4.5cm]{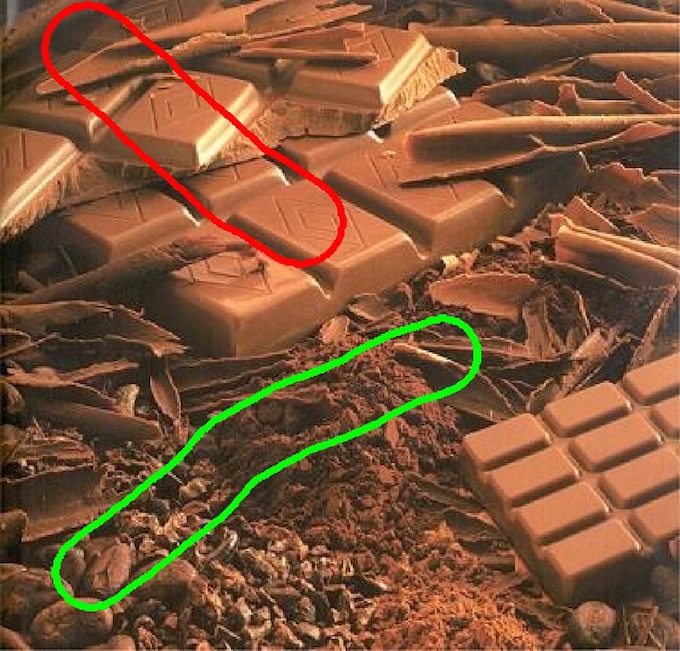} &
	\includegraphics[height=4.5cm]{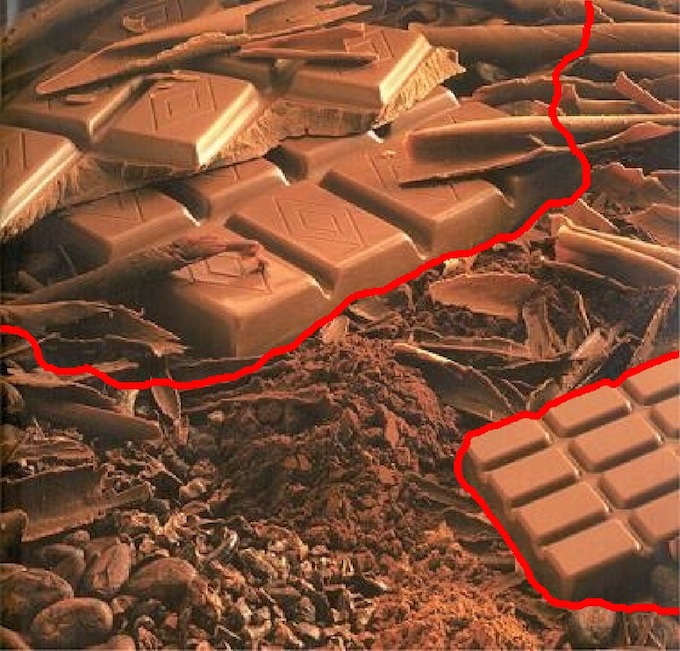} 
	\\
	Input & $\lambda = \infty$ 
	\end{tabular}
\caption{Texture segmentation using joint histograms of color gradient norms.
}
\label{fig:grad}
\end{figure}

In this experimental section, exemplar regions are defined by the user with scribbles (see Figures~\ref{fig:thres} to \ref{fig:grad}).
These regions are only used to built prior histograms, so erroneous labeling is tolerated.
Histograms $a$ and $b$ are built using hard-assignment on $M=8^n$ clusters, which are obtained with the K-means algorithm.
%\cmt{il faut détailler ?}

We use either RGB color features ($\F = \Id$ and $n=d=3$) or the gradient norm of color features ($\F = \lVert \nabla . \rVert$ and again $n=d=3$).
The cost matrix is defined from the Euclidean metric $\lVert \cdot \rVert$ in $\R^n$ space, combined with the concave function $1-e^{-\gamma{\lVert \cdot \rVert}{}}$, which is known to be more robust to outliers.
Region $R_t(u)$ is obtained with threshold $t=\tfrac{1}{2}$, as illustrated in Figure~\ref{fig:thres}.
% In our experiments, we chose $t=0.5$. 
Approximately $1$ minute is required to run $500$ iterations and segment 
a 1 Megapixel color image.
%an image of size $1200\times 900$.

\paragraph{Results}
Figure~\ref{fig:thres} shows the influence of the threshold $t$ used to get a binary segmentation.
A small comparison with the model of \cite{papa_aujol} is then given in Figures~\ref{fig:comp_l1} and \ref{fig:zeb}. This underlines the robustness of optimal transport distance with respect to bin-to-bin $L^1$ distance. Contrary to optimal transport, when a color is not present in the reference histograms, the $L^1$ distance does not take into account the color distance between bins which can lead to incorrect segmentation. The robustness is further illustrated in Figure~\ref{fig:zebras}. It is indeed possible to use a prior histogram from a different image, even with a different clustering of the feature space. Note that it is not possible with a bin-to-bin metric, which requires the same clustering.
%and can take bad decisions by associating some blue tones to the building area.
Figure~\ref{fig:comp_lambda} shows comparisons between the non-regularized model, quite fast but high dimensional model, with the regularized model, using a low dimensional formulation. One can see that setting a large value of $\lambda$ gives interesting results. 
%Namely, on the parrot example, the third parrot in the left part of the background is recovered. Moreover, the non-regularized model has wrongly segmented a leaf and this error is not present with the entropy  regularization. 
%When taking too low value of the regularization parameter,  ($\lambda=10$), the segmentation results get worse.
On the other hand, using a very small value of $\lambda$ always yields poor segmentation results.

Some last examples on texture segmentation are presented in Figure~\ref{fig:grad} 
where the proposed method is perfectly able to recover the textured areas.
We considered in this example the joint histograms of gradient norms on the $3$ color channels. 
Note that the complexity of the algorithm is the same as for color features, as long as we use the same number of clusters to quantize the feature space.

\section{Conclusion and future work}

Several formulations have been proposed in this work to incorporate transport-based distances in convex variational model for image processing, using either regularization of the optimal-transport or not.

Different perspectives have yet to be investigated, such as 
the final thresholding operation, the use of capacity transport constraint relaxation~\cite{Ferradans_siims14}, of other statistical features, of inertial and pre-conditionned optimization algorithms~\cite{Lorenz_Pock_inertial_FB_JMIV2015}, and the extension to region-based segmentation and to multi-phase segmentation problem.

%%%%%%%%%%%%%%%%%%%%%%%%%%%%%%%%%%%%%%%%%%%%%%%%%%%



\section*{Acknowledgments}
The authors would like to thanks Gabriel Peyré and Marco Cuturi for sharing their preliminary work
and Jalal Fadili for fruitful discussions on convex optimization.

% APPENDIX
%\clearpage
\appendix
\section{Proofs}%Appendices

\subsection{Proof of Corollary 1}%\ref{corollary:dual_simplexNmax}
\label{proof_coro}

\begin{proof}
Let us consider the problem:
{\small
\begin{equation}
\begin{split}
&\MK_{\lambda,\leq N}(a,b)\\ = &\min_{P \in {\Pp}(a,b)} \langle P,C \rangle +\frac{{1}}\lambda \langle P, \log (P/N) \rangle % {\color{red}\langle  P, \U \rangle}=N
+ {\iota_{\langle a,1\rangle = \langle b,1\rangle \leq N}(a,b)}.
\\
 %\hspace*{-10mm}
=&\min_{P } \max_{\substack{u ,\\v ,\\w \leq 0}} \hspace{-1pt} \left\{\hspace{-1pt}
\langle P,\frac{ \log P/N}\lambda\hspace{-1pt}+\hspace{-1pt} C\rangle 
\hspace{-1pt}+ \hspace{-1pt}\langle u,a - P \U_{M_b} \rangle\hspace{-1pt} +\hspace{-1pt}  \langle v,b - P^T \U_{M_a} \rangle\hspace{-1pt} + \hspace{-1pt}
%\right.\\
%&\qquad\qquad\qquad\qquad \left. 
 w (N-\U_{M_a}^T P \U_{M_b})\hspace{-1pt}\right\}\\ % {\color{red}\langle  P, \U \rangle}
  %\hspace*{-10mm}
 =&\max_{\substack{u ,\\v ,\\w \leq 0}}  \left\{ \langle u,a \rangle+ \langle v,b \rangle + Nw 
+ \min_{P}  \langle P,\frac{ \log P/N}\lambda +C - u \U_{M_b}^T - \U_{M_a} v^T - w\U_{M_a \times M_b}\rangle \right\} \\ 
\end{split}
\end{equation}
}

\noindent
where $\U_{M_a \times M_b} = \U_{M_a} \U_{M_b}^T$ is a matrix full of one, and
because $\langle u,P\U_{M_b} \rangle = \sum_{i=1}^{M_a} u_i \sum_{j=1}^{M_b} P_{i,j} =  \sum_{i,j} P_{i,j} u_{i} = \langle P, u \U_{M_b}^T \rangle $ and $\langle v,P^T\U_{M_a} \rangle = \langle P,  \U_{M_a} v^T \rangle$.

We then compute the first partial derivative of the Lagrangian 
$${\cal L}(P,u,v,w) := \langle P,\log P -\log N+ \lambda(C - u \U_{M_b}^T - \U_{M_a} v^T - w \U_{M_a \times M_b})
$$
$$
\partial_P {\cal L}(P^\star\hspace{-1pt},u,v,w) = \hspace{-1pt}\mathbf{0} \hspace{-1pt}= \log P^\star -\log N+ \lambda (C - u \U_{M_b}^T - \U_{M_a} v^T  - w \U_{M_a \times M_b}) + \U_{M_a \times M_b}
$$
so that:
\begin{equation}
\begin{split}
\log P_{i,j}^\star = -1 +\log N- \lambda(C_{i,j} - u_i - v_j - w)  \\
P_{i,j}^\star = e^{-1+\log N+\lambda w } \; e^{- \lambda(C_{i,j} - u_i - v_j)}
\end{split}
\end{equation}
Replacing this result back in the equation, we get:
{\small
\begin{equation}\label{calcul_intermediaire0}
\begin{split}
\MK_{\lambda,\leq N}(a,b) &=\max_{u ,v}\, \max_{w \leq 0} \left\{ \langle u,a \rangle+ \langle v,b \rangle + Nw
+ \frac1\lambda\langle P^\star,-\U_{M_a \times M_b}  \rangle \right\}
\\
& =\max_{u ,v}  \left\{ \langle u,a \rangle+ \langle v,b \rangle +\max_{w \leq 0}  Nw
- \frac{N}\lambda e^{-1+\lambda w} \sum_{i,j} e^{ - \lambda (C_{i,j} - u_i - v_j) } \right\}
\end{split}
\end{equation}
}

\paragraph{Case 1: $w<0$}
Let us first consider the case where the constraint is saturated, that is when $w<0$. 
When computing the maximum of $f(w) = Nw - \frac{N}\lambda e^{-1+\lambda w} K$ which is concave ($f''(w) < 0$), 
we obtain that 
$e^{-1+\lambda w^*} = \frac{1}{K} = {\left( \sum_{i,j} e^{- \lambda (C_{i,j} - u_i - v_j)}\right)}^{-1}$
so that the constraint $
\sum_{i,j} P_{i,j}^\star = N$ is checked.
Then, we have:
{\small
$$
f(w^*) =N w^* - \frac{N}\lambda
=  \frac{N}\lambda (\lambda w^*-1) = \frac{N}\lambda \log e^{\lambda w^* -1} = - \frac{N}\lambda\log \left(\sum_{i,j} e^{- \lambda(C_{i,j} - u_i - v_j)}\right)
$$
}
We finally get
\begin{equation}
\begin{split}
\MK_{\lambda,\leq N}(a,b)
& =\max_{u \in \R^{M_a},v \in \R^{M_b}}  \left\{ \langle u,a \rangle+ \langle v,b \rangle - \frac{ N}\lambda {  \log}\left(\sum_{i,j} e^{ - \lambda (C_{i,j} - u_i - v_j) }  \right)\right\}
\end{split}
\end{equation}
which means that $\MK_{\lambda,\leq N}(u,v)=\left(\frac{N}\lambda  {  \log}\left( \sum_{i,j} e^{ - \lambda (C_{i,j} - u_i - v_j) }  \right)\right)^*$.
As these functions are convex proper and lower semi-continuous, we have that $\MK^{**}_{\lambda,N}=\MK_{\lambda,N}$ which concludes the proof.

\paragraph{Case 2: $w=0$} Now we consider the case where the constraint is not saturated.
Going back to relation \eqref{calcul_intermediaire0}, the  unconstrained optimal value $w^*$ is still given by 
$$
	e^{\lambda w^*} = \frac{1}{\sum_{i,j} e^{-1 - \lambda (C_{i,j} - u_i - v_j)}}
$$
which involves 
$$
	w^*=-\frac{1}\lambda \log\left({\sum_{i,j} e^{-1- \lambda (C_{i,j} - u_i - v_j)}}\right).
$$
Hence, $w^*\ge 0$ as soon as ${\sum_{i,j} e^{1- \lambda (C_{i,j} - u_i - v_j)}} \le 1.$
In this case, the optimal value is therefore projected to $w^*=0$. Relation \eqref{calcul_intermediaire0} give us the following expression:
$$\max_{u \in \R^{M_a},v \in \R^{M_b}}  \left\{ \langle u,p \rangle+ \langle v,q \rangle 
- \frac{N}\lambda \sum_{i,j} e^{ -1 - \lambda (C_{i,j} - u_i - v_j) } \right\},$$
which concludes the proof.

\end{proof}

\subsection{Proof of proposition 2}
\label{sec:proof_lipschitz_dual}

\begin{proof}
The derivative $\nabla \MK^{*}_{\lambda} (X) $ with $X=(u,v)$  is lipschitz continuous iff there exist $L_{\MK^*}>0$ such that 
$$\lVert  \nabla \MK^{*}_{\lambda}(X)-\nabla \MK^{*}_{\lambda}(X') \rVert\leq L_{\MK^*}\lVert X-X'\rVert.$$
We denote as $U_1$ the set of vectors $X=(u,v)$ such that $\sum_{i,j}e^{-1-\lambda (C_{i,j} -u_i - v_j)}\geq 1$ and $V_1$ the set $\sum_{i,j}e^{-1-\lambda (C_{i,j} -u_i - v_j)}\leq 1$.
We will detail three cases. 
\paragraph{Case 1}We first consider $X,X'\in U_1$.
As  $\nabla \MK^{*}_{\lambda} (X) $ is derivable in the set $U_1$, it is a lipschitz function iff the norm of the Hessian matrix $\H$ of $\MK^{*}_{\lambda} (u,v) $ is bounded. Being $\{\mu_i\}_{i=1}^{M_a+M_b}$ the eigenvalues of $\H$, its $l^2$ norm is defined as $\lVert\H\rVert=\max_i |\mu_i|$. 
Moreover, as $\MK^{*}_{\lambda}$ is convex, we know that all its eigenvalues are non negative.
Thus, we have that the norm of $\H$ is bounded by its trace: $\lVert\H\rVert\leq Tr(\H)=\sum_i \mu_i=\sum_i \H_{ii}$.

The Hessian matrix $\H$ of   $\MK^{*}_{\lambda} (x,y)$ is defined as:
$$\H=\begin{bmatrix}\H^{11}&\H^{12}\\(\H^{12})^T&\H^{22}\end{bmatrix},$$
with $ \H^{11}=\nabla_1 \nabla_1 \MK^{*}_{\lambda}$, $\H^{12}=\nabla_2\nabla_1  \MK^{*}_{\lambda}$ and $\H^{22}=\nabla_2\nabla_2 \MK^{*}_{\lambda}$.
One can first observe that 
$$
(\nabla_1 \MK^{*}_{\lambda}(u,v))_i
=\partial_{u_i}\MK^{*}_{\lambda}(u,v)
= { N} P^\star(u_i,v) \U_{M_b}
= {N} \frac{\sum_l  e^{- \lambda (C_{i,l} - u_i - v_l) }}{\sum_{k,l}  e^{- \lambda (C_{k,l} - u_k^{} - v_l) }}.
$$
Hence the diagonal element of the matrix  are $\H^{11}_{ii}=\partial^2_{u_i} \MK^{*}_{\lambda}(u,v)$
and $\H^{22}_{jj}=\partial^2_{v_j} \MK^{*}_{\lambda}(x,y)$. They read:
\begin{equation}
\begin{split}
\H^{11}_{ii}&=\lambda {N} \frac{\sum_l e^{- \lambda (C_{i,l} - u_i-v_l) }{\sum_{k\neq i,l}  e^{- \lambda (C_{k,l} - u_k^{} -v_l) }}}{ {\left(\sum_{k,l}  e^{- \lambda (C_{k,l} - u_k^{}- v_l) } \right)}^2 }\\
\H^{22}_{jj}&=\lambda {N} \frac{\sum_k^{} e^{- \lambda (C_{k,j} - u_k^{}-v_j) }{\sum_{k,l\neq j}  e^{- \lambda (C_{k,l} - u_k^{} -v_l) }}}{ {\left(\sum_{k,l}  e^{- \lambda (C_{k,l} -u_k^{}- v_l) } \right)}^2 }\\
\end{split}
\end{equation}
Computing the trace of the matrix $\H$, we have:
$$\lVert\H\rVert\leq Tr(\H)=\sum_i \H^{11}_{ii}+\sum_j\H^{22}_{jj}\leq 2\lambda N.$$

\paragraph{Case 2} We now consider $X,X'\in V_1$.
In this case, we have for $X=(u,v)$:
$$\MK^*_{\lambda} (u,v) =\frac1\lambda \sum_{i,j} e^{-1-\lambda (C_{i,j} -u_i - v_j)}.$$
As the double derivative w.r.t $u_i$ is $\partial_{u_i}^2\MK^*_{\lambda} (u,v)=\lambda\sum_j e^{-1-\lambda (C_{i,j} -u_i - v_j)}$,
the trace of the Hessian matrix reads: 
$$Tr(\H)=\lambda \left(\sum_i\sum_j e^{-1-\lambda (C_{i,j} -u_i - v_j)}+\sum_j\sum_i e^{-1-\lambda (C_{i,j} -u_i - v_j)}\right)\leq 2\lambda N,$$
since $(u,v)\in V_1$.
\paragraph{Case 3} We consider $X\in U_1$ and $X'\in V_1$. We denote as $Y$ a vector that lies in the segment $[X; X']$ and belongs to the boundary of $U_1$ and $V_1$. We thus have 
$\lVert X-Y \rVert+\lVert X'-Y\rVert=\lVert X-X'\rVert$ so that
{\small
\begin{equation}
\begin{split}
&\lVert \nabla \MK^{*}_{\lambda}(X) - \nabla \MK^{*}_{\lambda}(X') \rVert
\\
& \quad \leq  \lVert \nabla \MK^{*}_{\lambda}(X)-\nabla \MK^{*}_{\lambda}(Y)\rVert + \lVert \nabla \MK^{*}_{\lambda}(X')-\nabla \MK^{*}_{\lambda}(Y)\rVert 
\\
& \quad \leq 2\lambda N (\lVert X-Y\rVert + \lVert X'-Y\rVert)=2\lambda N\lVert |X-X'\rVert
\\
\end{split}
\end{equation}
}

\end{proof}

\subsection{Proof of proposition 3}
\label{proof:prox_bidual}
\begin{proof}
We are interested in the proximity operator of 
$
	f^*(q) = \frac{N}{\lambda} \dotp{e^{\lambda (q-c) - \U}}{\U}
$.
First notice that the proximity operator of $f^*$ can be computed easily from the proximity operator of $f$ through Moreau's identity:
$$\text{prox}_{\tau f} (p)= p-\tau \text{prox}_{ f^*/\tau} (p/\tau).$$
We now recall that the Lambert function $W$ is defined as:
$$
z = we^w \Leftrightarrow w = W(z)
$$
where $w$ can take two real values for $z\in ]-\frac1e,0]$, and only one on $]0,\infty[$, as illustrated in Figure \ref{fig:lambert}.
As $z$ will always be positive in the following, we do not consider complex values.

\begin{figure}[!htb]
	\centering
	\includegraphics[width=0.65\textwidth]{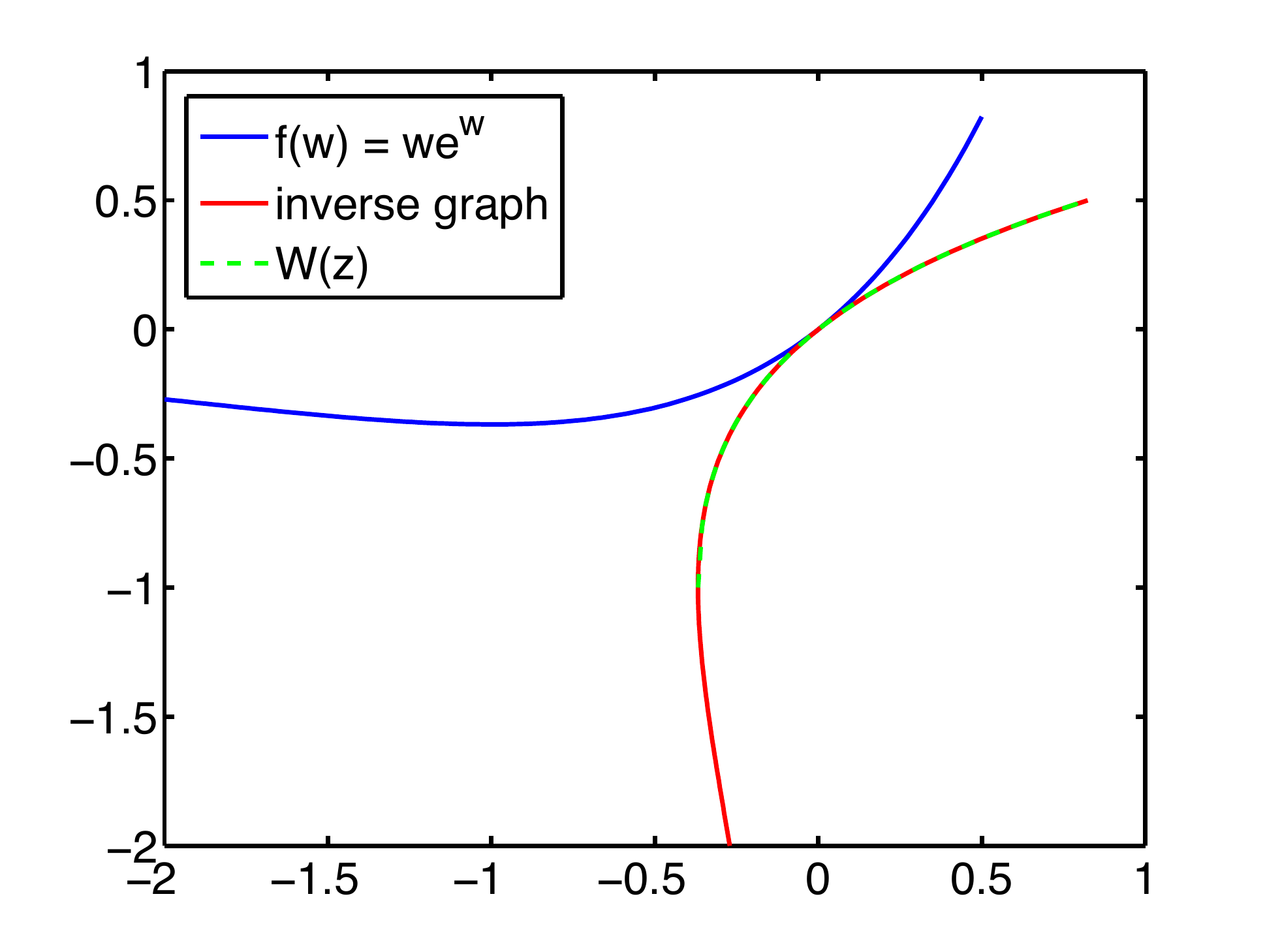}
	\caption{Lambert function $W(z)$.}
	\label{fig:lambert}
\end{figure}

We recall that the proximity operator of $f^*$ at point $p$ reads:
\begin{equation}
\begin{split}
\text{prox}_{\tau f^*}(p)&=\uargmin{q} \frac{||q-p||^2}{2\tau}+f^*(q)\\
&=\uargmin{q} \sum_k^{}\frac{||q_k^{}-p_k^{}||^2}{2\tau}+\frac{N}{\lambda} \sum_k^{} e^{\lambda(q_k^{} - c_{k})-1}=q^\star
\end{split}
\end{equation}
This problem is separable and can be done independently $\forall k\in [1,M_a^{}M_b^{}]$.
Deriving the previous relation with respect to $q_k^{}$, the optimality condition reads:
\begin{equation}
\begin{split}
&q^\star_k-p_k^{}+\tau N e^{\lambda(q^\star_k - c_{k})-1}=0
\\
\Leftrightarrow&(p_k^{}-q_k^*)e^{-\lambda q^\star_k}=\tau N e^{-\lambda c_{k}-1}
\\
\Leftrightarrow&\lambda(p_k^{}-q_k^*)e^{\lambda (p_k^{}-q_k^*)}
= \lambda\tau N e^{\lambda (p_k^{}-c_{k})-1}.
\end{split}
\end{equation}
Using Lambert function, we get:
\begin{equation}
\begin{split}
&\lambda(p_k^{}-q^\star_k)=W(\lambda\tau N e^{\lambda (p_k^{}-c_{k})-1})\\
\Leftrightarrow&q_k^*=p_k^{}-\frac1\lambda W(\lambda\tau N e^{\lambda (p_k^{}-c_{k})-1}).
\end{split}
\end{equation}
As $f^*/\tau$ is convex, the $\text{prox}$ operator is univalued, then only one possible value of the Lambert function is admissible, which is indeed the case since $z$ is strictly positive).
The proximity operator of $\tau f^*$ thus reads
\begin{equation}
\text{prox}_{\tau f^*}(p) = p - \frac{1}{\lambda} W\left(  \lambda \tau N e^{\lambda (p-c) - 1}  \right).
\end{equation}
which is in agreement with \cite{CombPesq} (Chapter 10, page 190, property xii).
\end{proof}

% ------------- BIBLIO -------------------

\bibliographystyle{abbrv}
%\bibliography{refs}

\end{document}